\documentclass[10pt,journal,compsoc]{IEEEtran}
\usepackage{amsmath,amsfonts}
\usepackage{algorithmic}
\usepackage{algorithm}
\usepackage{array}
\usepackage{textcomp}
\usepackage{stfloats}
\usepackage{url}
\usepackage{verbatim}
\usepackage{cite}
\usepackage{amssymb}
\usepackage{amsmath}
\usepackage{multirow}
\usepackage{booktabs}
\usepackage{graphicx} 
\usepackage{subcaption} 
\usepackage{booktabs}  
\usepackage{multirow}  
\usepackage{threeparttable}  
\usepackage[table]{xcolor}  
\usepackage{booktabs}      
\usepackage{multirow}       
\usepackage{threeparttable} 
\usepackage[table]{xcolor}

\begin{document}

\title{SMART-Ship: A Comprehensive Synchronized Multi-modal Aligned Remote Sensing Targets Dataset and Benchmark for Berthed Ships Analysis}

\author{
Chen-Chen Fan*, Peiyao Guo*, Linping Zhang, Kehan Qi, Haolin Huang, Yong-Qiang Mao \\  Yuxi Suo, Zhizhuo Jiang, Yu Liu, You He
\thanks{
C.-C. Fan, P. Guo, L. Zhang, Y.-Q. Mao, Y. Suo, Y. Liu, and Y. He are with the Department of Electronic Engineering, Tsinghua University, Beijing, China. K. Qi, H. Huang, and Z. Jiang are with Tsinghua Shenzhen International Graduate School, Shenzhen, China. (E-mail: fanchenchen@mail.tsinghua.edu.cn, guopeiyao@mail.tsinghua.edu.cn)
}
\thanks{*These authors contributed equally to this work. Corresponding author: Zhizhuo Jiang, Yu Liu (E-mail: jiangzhizhuo@sz.tsinghua.edu.cn, liuyu\_thu@mail.tsinghua.edu.cn).}
\thanks{This work was supported in part by the National Natural Science Foundation of China under Grant 62425117, 62401335, and 62293544, and in part by China Postdoctoral Science Foundation under Grant 2024T170493, 2025M773474, 2025M773484, and in part by the Postdoctoral Innovation Talents Support Program under Grant BX20240178, BX20250413.}
}

\markboth{IEEE TRANSACTIONS ON PATTERN ANALYSIS AND MACHINE INTELLIGENCE}%
{Shell \MakeLowercase{\textit{et al.}}: A Sample Article Using IEEEtran.cls for IEEE Journals}



\IEEEtitleabstractindextext{
\begin{abstract}
Given the limitations of satellite orbits and imaging conditions, multi-modal remote sensing (RS) data is crucial in enabling long-term earth observation.
However, maritime surveillance remains challenging due to the complexity of multi-scale targets and the dynamic environments. 
To bridge this critical gap, we propose a \textbf{S}ynchronized \textbf{M}ulti-modal \textbf{A}ligned \textbf{R}emote sensing \textbf{T}argets dataset for berthed \textbf{ship}s analysis (\textbf{SMART-Ship}), containing spatiotemporal registered images with fine-grained annotation for maritime targets from five modalities: visible-light, synthetic aperture radar (SAR), panchromatic, multi-spectral, and near-infrared. 
Specifically, our dataset consists of 1092 multi-modal image sets, covering 38,838 ships. Each image set is acquired within one week and registered to ensure spatiotemporal consistency. Ship instances in each set are annotated with polygonal location information, fine-grained categories, instance-level identifiers, and change region masks, organized hierarchically to support diverse multi-modal RS tasks. 
Furthermore, we define standardized benchmarks on five fundamental tasks and comprehensively compare representative methods across the dataset. Thorough experiment evaluations validate that the proposed SMART-ship dataset could support various multi-modal RS interpretation tasks and reveal the promising directions for further exploration.
\end{abstract}
}

\maketitle
\IEEEdisplaynontitleabstractindextext

\section{Introduction}
\label{sec:introduction}

\begin{figure*}
    \centering
    \includegraphics[width=0.95\linewidth]{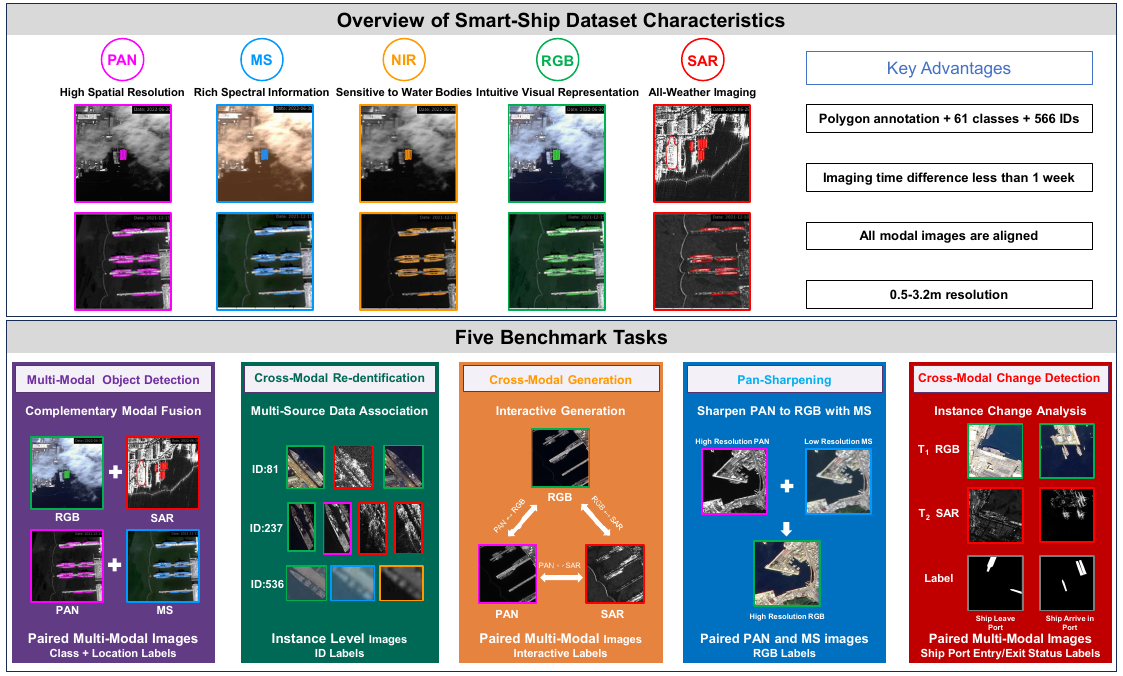}
    \caption{
    The proposed SMART-Ship dataset. This comprehensive multi-modal maritime dataset features precise polygon annotations, instance-level identification across 566 vessels, multi-scale spatial coverage, and temporal alignment, supporting five benchmark tasks for multi-modal image analysis and evaluation.
    }
    \label{fig_001}
\end{figure*}

\IEEEPARstart{R}ecently, remote sensing (RS) technology with multiple sensors has attracted more attention for long-term society development, e.g., urban planning, environment monitoring, transportation management, and maritime surveillance~\cite{STAR, HBDNet, FarSeg++, YOLOFA, MetaEarth, gu2020semi}. 
As diverse sensors offer heterogeneous properties of the scenes, numerous explorations~\cite{jin2022lagconv, Ciotola2022,deng2020detail,razakarivony2016vehicle,sun2022drone,qingyun2022, ban,chen2021bit,cf,li2022asymmetric,SHEN2024109913,zhang2024cooperative, WflmGAN2022,SemiD2022,cai2024proxy,wang2024hgrmc,gao2025hoss} leverage complementary properties between multi-modal data to improve the accuracy and robustness of RS interpretation performance. For instance, pan-sharpening tasks ~\cite{jin2022lagconv, Ciotola2022,deng2020detail} combine the high-definition structure details from panchromatic (PAN) imagery with spectral cues of multi-spectral (MS) data to obtain high-quality scene observation. Compared with single-modal data, leveraging multi-modal data can also provide discriminative clues
for Object Detection~\cite{razakarivony2016vehicle, sun2022drone,qingyun2022}, Change Detection~\cite{ban,chen2021bit,cf}. 
Recent works ~\cite{li2022asymmetric,SHEN2024109913,zhang2024cooperative} leverage the imaging of Synthetic Aperture Radar (SAR) to compensate for occlusion or saturation in optical imagery, even synthesize visible-light (RGB) features from the SAR modality~\cite{WflmGAN2022,SemiD2022}, thus facilitating robust object detection, object retrieval/re-identification~\cite{cai2024proxy,wang2024hgrmc,gao2025hoss} and continuous monitoring under adverse conditions such as cloud cover.

Comprehensive long-term earth observation requires the integration of multiple RS interpretation tasks. However, existing multi-modal RS datasets are often designed for specific tasks and scenarios.
Usually, current multi-modal datasets for object detection~\cite{chen2023mmship, liu2023mssd} consist of RGB–NIR image pairs with coarse category labels and rectangular bounding boxes. They provide few instance-level annotations such as precise ship contours or fine-grained classifications, which are insufficient for other tasks like object re-identification.
Datasets for cross-modal generation~\cite{shermeyer2020spacenet,huang2021qxs,wang2019sar} and pan-sharpening~\cite{qb,wv2,wv3} generally offer spatially and temporally aligned image pairs (e.g., RGB-SAR or PAN-MS), while those for change detection~\cite{cali,tou,subpixel} include dual-temporal observations from the same region to grasp the scene dynamics. 
Despite their contributions, these datasets primarily focus on pixel-level scene analysis and offer limited support for object-level semantic understanding.
The lack of rich, fine-grained annotations significantly restricts the interaction between different RS tasks, thereby hindering comprehensive multi-modal scene interpretation.
Furthermore, most datasets center on the urban or land scenarios with repetitive structures, which differ greatly from maritime scenes. The maritime scene interpretation faces unique challenges due to the diversity in ship categories, scales, and dynamic imaging conditions. It is also challenging to offer both spatially and temporally registered image pairs of maritime targets within short intervals (e.g., one week), which is essential for capturing scene dynamics. Hence, there is an urgent need to collect multi‑source RS data that covers diverse ships and dynamic environments for various maritime downstream tasks.

To further explore the multi-modal RS analysis in maritime scenes, we present the \textbf{S}ynchronized \textbf{M}ulti-modal \textbf{A}ligned \textbf{R}emote sensing \textbf{T}argets for \textbf{Ships} (\textbf{SMART-Ship}) dataset. SMART-Ship is the first dataset to provide a unified resource with static shores and dynamic ships features for the diverse maritime scene interpretation. It offers a rich and diverse collection of data that enables the research community to explore the frontiers of multi-modal learning. The contributions of our work are summarized as follows:

\begin{itemize}
\item To the best of our knowledge, we are the first to propose a multi-modal ship dataset with fine-grained annotations for diverse remote sensing tasks. The dataset comprises 1092 registered scenes captured by multiple satellites within one week, covering five complementary modalities (i.e., visible-light, SAR, panchromatic, multi-spectral, and near-infrared). It spans a wide range of ships varying in shape, localization, scale, and category, with hierarchical fine-grained annotations, which support multiple RS interpretation tasks. 
\item We define five benchmark tasks upon the proposed dataset, including detection, re-identification, cross-modal generation, pan-sharpening, and change detection. Typical solutions for each multi-modal analysis task are comprehensively evaluated on our benchmark to validate the effectiveness of our dataset. 
\item We further analyze the benchmark results across multiple downstream tasks, revealing the challenges
and opportunities for multi-modal fusion, cross-modal learning, or multi-task collaboration in maritime RS interpretation.
\end{itemize}

This paper is organized as follows. Section~\ref{sec:related} reviews Related work on multi-modal RS datasets and corresponding downstream tasks. Section ~\ref{sec:dataset} introduces the details of the SMART-Ship dataset, including its acquisition, composition, and annotation characteristics. In section~\ref{sec:benchmark}, we establish the benchmarks for five tasks and provide a comprehensive evaluation of representative methods.
Finally, we discuss potential directions for future research upon the dataset in section~\ref{sec:analysis} and make the conclusion in section~\ref{sec:conclusion}.

\section{Related Work} 
\label{sec:related}

\begin{table*}[t]
\centering
\caption{Comparison of Multi-Modal Remote Sensing Datasets.}
\label{tab:datasets}
\resizebox{1\textwidth}{!}{
\begin{tabular}{p{2.7cm}|c|p{1.5cm}|c|c|p{0.4cm}p{0.4cm}p{0.4cm}p{0.4cm}p{0.4cm}|p{0.3cm}p{0.35cm}p{0.3cm}p{0.45cm}p{0.3cm}|p{0.5cm}p{0.3cm}p{0.4cm}}
\toprule
\textbf{Dataset} 
  & \textbf{Year} 
  & \textbf{Number} 
  & \textbf{Image Size} 
  & \textbf{Resolution} 
  & \textbf{RGB} 
  & \textbf{SAR} 
  & \textbf{PAN} 
  & \textbf{MS} 
  & \textbf{NIR} 
  & \textbf{Det} 
  & \textbf{ReID} 
  & \textbf{Gen} 
  & \textbf{Sharp} 
  & \textbf{CD} 
  & \textbf{BBox} 
  & \textbf{Cls} 
  & \textbf{Inst} \\
\midrule
VEDAI~\cite{razakarivony2016vehicle}  
  & 2016 & $2\times1,200$   & $1,024\times1,024$ & $0.125$m 
  & \checkmark & $\times$ & $\times$ & $\times$ & \checkmark 
  & \checkmark & $\times$ & $\times$ & $\times$ & $\times$ 
  & HBB & 9 & -- \\
DroneVehicle~\cite{sun2022drone}  
  & 2022 & $2\times28,439$ & $640\times512$ & --\textsuperscript{\dag} 
  & \checkmark & $\times$ & $\times$ & $\times$ & \checkmark 
  & \checkmark & $\times$ & $\times$ & $\times$ & $\times$ 
  & OBB & 5 & -- \\
MMShip~\cite{chen2023mmship}  
  & 2023 & $2\times5,016$  & $512\times512$ & $10$m 
  & \checkmark & $\times$ & $\times$ & $\times$ & \checkmark 
  & \checkmark & $\times$ & $\times$ & $\times$ & $\times$ 
  & HBB & 1 & -- \\
MSSD~\cite{liu2023mssd}  
  & 2024 & $2\times2,494$   & $512\times512$ & $10$m  
  & \checkmark & $\times$ & $\times$ & $\times$ & \checkmark 
  & \checkmark & $\times$ & $\times$ & $\times$ & $\times$ 
  & HBB & 1 & -- \\
\midrule
HOSS ReID~\cite{gao2025hoss}  
  & 2025 & $1065+767$   & --\textsuperscript{\ddag} & 0.75-1m 
  & \checkmark & \checkmark & $\times$ & $\times$ & $\times$ 
  & $\times$ & \checkmark & $\times$ & $\times$ & $\times$ 
  & HBB & 1 & 449 \\
\midrule

SEN1-2~\cite{schmitt2018sen1}  
  & 2018 & $2\times282,384$ & $600\times600$ & 5-10m 
  & \checkmark & \checkmark & $\times$ & $\times$ & $\times$ 
  & $\times$ & $\times$ & \checkmark & $\times$ & $\times$ 
  & -- & -- & -- \\
WHU-SEN-City~\cite{wang2019sar}  
  & 2019 & $2\times18,542$ & $256\times256$ & 10-22m
  & \checkmark & \checkmark & $\times$ & $\times$ & $\times$ 
  & $\times$ & $\times$ & \checkmark & $\times$ & $\times$ 
  & -- & -- & -- \\
MSAW~\cite{shermeyer2020spacenet}  
  & 2020 & $2\times3,401$   & $900\times900$ & 0.5m 
  & \checkmark & \checkmark & $\times$ & $\times$ & $\times$ 
  & $\times$ & $\times$ & \checkmark & $\times$ & $\times$ 
  & -- & -- & -- \\
QXS-SAROPT~\cite{huang2021qxs}  
  & 2021 & $2\times20,000$ & $256\times256$ & 1m 
  & \checkmark & \checkmark & $\times$ & $\times$ & $\times$ 
  & $\times$ & $\times$ & \checkmark & $\times$ & $\times$ 
  & -- & -- & -- \\
SAR2Opt~\cite{zhao2022comparative}  
  & 2022 & $2\times2,076$ & $600\times600$ & 1m 
  & \checkmark & \checkmark & $\times$ & $\times$ & $\times$ 
  & $\times$ & $\times$ & \checkmark & $\times$ & $\times$ 
  & -- & -- & -- \\
SMILE~\cite{xia2024sossf}  
  & 2024 & $3\times6,177$ & $512\times512$ & 30m 
  & $\times$ & \checkmark & $\times$ & \checkmark & \checkmark 
  & $\times$ & $\times$ & \checkmark & $\times$ & $\times$ 
  & -- & -- & -- \\
\midrule
QuickBird~\cite{qb} 
  & 2013 & $3\times19044$ & $256\times256$ & 0.61m, 2.44m 
  & \checkmark & $\times$ & \checkmark & \checkmark & $\times$ 
  & $\times$ & $\times$ & $\times$ & \checkmark & $\times$
  & -- & -- & -- \\
WorldView3~\cite{wv3}
  & 2014 & $3\times10794$ & $256\times256$ & 0.3m, 1.2m
  & \checkmark & $\times$ & \checkmark & \checkmark & $\times$ 
  & $\times$ & $\times$ & $\times$ &\checkmark & $\times$
  & -- & -- & -- \\
GaoFen2~\cite{gf2}
  & 2017 & $3\times22010$ & $256\times256$ & 1m, 4m
  & \checkmark & $\times$ & \checkmark & \checkmark & $\times$ 
  & $\times$ & $\times$ & $\times$ & \checkmark & $\times$ 
  & -- & -- & -- \\
EvalSet~\cite{meng2020large}
  & 2020 & $3\times2270$ & $1024\times1024$ & 0.3-2m, 1.24-8m
  & \checkmark & $\times$ & \checkmark & \checkmark & $\times$ 
  & $\times$ & $\times$ & $\times$ & \checkmark & $\times$ 
  & -- & -- & -- \\
\midrule
Sulzberger~\cite{sul}
  & 2011 & $2\times1^{*}$  & $256\times256$ & --\textsuperscript{\dag} 
  & $\times$ & \checkmark & $\times$ & $\times$ & $\times$ 
  & $\times$ & $\times$ & $\times$ & $\times$ & \checkmark 
  & -- & -- & -- \\ 
Californi~\cite{cali}  
  & 2019 & $2\times1^{*}$  & $3,500\times2,000$ & 15m 
  & \checkmark & \checkmark & $\times$ & $\times$ & $\times$ 
  & $\times$ & $\times$ & $\times$ & $\times$ & \checkmark 
  & -- & -- & -- \\ 
Toulouse~\cite{tou}
  & 2020 & $2\times1^{*}$  & $2,604\times4,404$ & 2m 
   & \checkmark & \checkmark & $\times$ & $\times$ & $\times$ 
  & $\times$ & $\times$ & $\times$ & $\times$ & \checkmark 
  & -- & -- & -- \\
Wuhan ~\cite{whucd}
  & 2022 & $2\times1^{*}$  & $11,216\times13,694$ & 3m 
  & \checkmark & \checkmark & $\times$ & $\times$ & $\times$ 
  & $\times$ & $\times$ & $\times$ & $\times$ & \checkmark 
  & -- & -- & -- \\ 
\midrule
\textbf{SMART-Ship}  
  & 2025 & $5\times\,1,092$ & $1,024\times1,024$ & 0.5-3.2m 
  & \checkmark & \checkmark & \checkmark & \checkmark & \checkmark 
  & \checkmark & \checkmark & \checkmark & \checkmark & \checkmark 
  & Poly & 61 & 566 \\
\bottomrule
\end{tabular}
}

\vspace{5pt}
\begin{minipage}{\textwidth}
\footnotesize
Note: 
\textsuperscript{\dag}Resolution unspecified;
\textsuperscript{\ddag}all images are object chips with varying sizes;
{*}paired wide-swath image.
Det: object detection; 
ReID: object re-identification; 
Gen: cross-modal generation; 
Sharp: pan-sharpening; 
CD: change detection; 
BBox: bounding box type;
HBB: horizontal bounding box;
OBB: oriented bounding box;
Poly: polygon bounding box;
Cls: number of classes; 
Inst: number of instance-level labels.
\end{minipage}
\end{table*}

\subsection{Multi-Modal Remote Sensing Datasets}

In recent years, the release of multi-modal RS datasets has significantly promoted the development of cross-modal RS data fusion research. Representative modalities include PAN, MS, NIR, SAR, and RGB. These modalities offer distinct advantages and limitations in terms of spatial resolution, spectral richness, and resilience to environmental conditions. However, most existing multi‑modal RS datasets are tailored to individual tasks. A comprehensive multi‑modal dataset, therefore, would be a keystone for advancing versatile, task‑agnostic RS fusion research.

\textbf{Object Detection Dataset.} Multi-modal RS datasets for object detection usually consist of RGB and NIR modalities at the same spatial resolution. Limited by the challenge of obtaining multi-satellite data, existing multi-modal RS datasets are mainly RGB and NIR images. VEDAI \cite{razakarivony2016vehicle} represents an early multi-modal RS dataset comprising 1,200 pairs of RGB and NIR images, establishing a benchmark for MS vehicle detection tasks. Later, DroneVehicle~\cite{sun2022drone} provides a vehicle detection dataset acquired under varying illumination conditions. MMShip \cite{chen2023mmship} and MSSD \cite{liu2023mssd} are proposed for ship detection, which advances research in multi-modal maritime monitoring. Though the above datasets have driven remarkable progress in RS object detection, they still exhibit some shortcomings, including limited annotation granularity and insufficient modality diversity.

\textbf{Re-ID Dataset.} Recently, Wang \textit{et al.} introduce HOSS ReID~\cite{gao2025hoss}, the first cross-modal remote sensing ship Re-ID dataset, which provides RGB and SAR images annotated with identities. However, HOSS ReID~\cite{gao2025hoss} remains limited in two critical aspects. First, it only supports two modalities (RGB and SAR), thereby constraining the scope of cross-modal relationship modeling and precluding analysis of spectral complementarity among other widely used sensors. Second, its annotation structure focuses solely on identity-level labels without incorporating additional geometric or semantic information. In contrast, the SMART-Ship dataset provides more detailed category annotations and target neighborhood information, which is expected to support future research.

\textbf{Cross‐modal Generation Dataset.} RS image cross‐modal generation plays a pivotal role in aligning feature representations between modalities, thereby facilitating complementary information exchange and boosting downstream interpretation tasks. Several large‐scale generation datasets have been released to foster this line of research. 
The MSAW dataset~\cite{shermeyer2020spacenet} provides SAR-RGB images over a 120 $ \text{km}^2 $ area. The QXS‐SAROPT dataset~\cite{huang2021qxs} includes 20,000 SAR-RGB pairs from GaoFen‐3 and Google Earth, covering Qingdao, Shanghai, and San Diego. The WHU‐SEN‐City dataset~\cite{wang2019sar} collects SAR-RGB pairs from multiple satellites across 32 Chinese cities. The SEN1‐2 dataset~\cite{schmitt2018sen1} contains 282,384 paired patches of Sentinel‐1 SAR and Sentinel‐2 optical imagery. The SAR2Opt dataset~\cite{zhao2022comparative} offers high-resolution TerraSAR-X SAR and optical image pairs. Finally, the SMILE dataset~\cite{xia2024sossf} is the first global collection of Sentinel-1, MODIS, and Landsat-8 triplets, enabling spatial–spectral fusion.
While these existing datasets significantly contribute to cross-modal generation research, they are primarily limited to dual-modal configurations and general land-use scenarios. In contrast, SMART-Ship enables multi-directional generation across five modalities, supporting more diverse application scenarios than traditional dual-modal datasets. The dataset's high spatial resolution (0.5–3.2m) and standardized format provide favorable conditions for high-quality generation and stable training. Moreover, the polygon-level annotations have the potential to support fine-grained generation control for precise cross-modal ship synthesis. As the first ship-focused multi-modal dataset, SMART-Ship alleviates the gap in port cross-modal synthesis applications.

\textbf{Pan-Sharpening Dataset.}  In the pan-sharpening domain, typical RS datasets include WorldView-3~~\cite{wv3} (high resolution with 10,794 pairs covering Rio and Tripoli), GaoFen-2~\cite{gf2} (moderate resolution with extensive coverage in Guangzhou), and QuickBird ~\cite{qb}(lower resolution but large volume for Indianapolis), which mainly focus on terrestrial urban environments. Besides these mentioned training datasets, an evaluation dataset is also included in EvalSet~\cite{meng2020large}. However, these datasets lack diverse maritime and port scenarios, which present unique challenges due to complex spectral characteristics, including variations in water surface reflectance, complexity of port infrastructure, and different types of ships. The SMART-Ship dataset sufficient this shortage by providing marine and port scenarios with complex spectral characteristics of water bodies and port infrastructure. The SMART-Ship dataset is an important supplement in pan-sharpening research, enabling more comprehensive algorithm development across diverse environmental contexts.

\textbf{Change Detection Dataset.} 
Several existing cross-modal change detection datasets have contributed to research development, though each has distinct limitations for maritime applications. The Sulzberger dataset provides single-modality data for glacier change detection\cite{sul}, while the California\cite{cali}, Toulouse\cite{tou}, and Wuhan\cite{whucd} datasets combine optical and SAR imagery but operate with ultra-wide field of view at relatively low ground resolution (10-30m) due to satellite-based observation requirements. These existing datasets are predominantly constrained to scene-level image pairs with insufficient ground resolution for detailed maritime object detection, indicating the need for multi-modal aligned datasets with high ground resolution specifically designed for maritime scenarios.
The SMART-Ship dataset is expected to address these limitations by providing 1,092 pairs of high-resolution registered RGB and SAR image pairs specifically tailored for marine and port environments. Unlike existing datasets, this dataset employs instance-level change annotations with enhanced ground resolution, which is expected to enable precise vessel change detection and monitoring of detailed ship movements including port entry and departure activities. This contribution is expected to alleviate the limitations in cross-modal change detection research for maritime applications.

\textbf{Advantages of SMART-Ship Dataset.} 
Table~\ref{tab:datasets} summarizes the characteristics of existing multi-modal RS datasets alongside our proposed SMART-Ship. 
In response to the fragmented and task-specific limitations of previous datasets, SMART-Ship is designed as a unified benchmark that simultaneously addresses multiple key challenges across diverse RS applications.
For object detection, SMART-Ship provides high-precision polygon-level annotations for 38,838 ships, enabling fine-grained localization beyond coarse bounding boxes. In cross-modal Re-ID, it is the first dataset to offer instance-level identity labels across five modalities, thus facilitating fine-grained retrieval across heterogeneous domains. For cross-modal generation, the availability of registered five-modality imagery captured within one week empowers research on consistent and structurally meaningful cross-modal generation. In the domain of pan-sharpening, SMART-Ship is the first set of precisely registered PAN and MS image pairs specifically tailored for maritime and port environments. This unique focus strengthens the weakness of dedicated pan-sharpening resources for ship and port scenarios.
For cross-modal change detection, SMART-Ship provides 1,092 pairs of meticulously registered 1,024×1,024 high-resolution RGB and SAR images with comprehensive instance-level change annotations, which is expected to promote cross-modal change detection algorithms to focus more on target instance variations. With its rich modality diversity, detailed annotations, and broad task compatibility, SMART-Ship fills a critical gap in the current landscape of multi-modal RS datasets and lays a solid foundation for advancing unified multi-task, cross-modal learning frameworks.

\subsection{Multi-Modal Solutions for Downstream Tasks}
\textbf{Multi‐Modal Ship Detection Methods.}
Multi‐modal ship detection enhances robustness and accuracy by exploiting complementary information from multi-sensor data. Early data-level fusion, typically implemented via channel-wise concatenation or pixel-wise addition of registered inputs, offers computational simplicity but is highly sensitive to misalignment~\cite{qingyun2022}. To alleviate this drawback, feature‐level fusion first encodes each modality independently and then merges representations, e.g., by element‐wise addition or channel concatenation~\cite{qingyun2022}, or via attention and Transformer modules that adaptively learn inter‐modal correlations~\cite{qingyun2021,SHEN2024109913}. Decision‐level fusion works by ensembling modality‐specific detectors, as in the Bayesian framework ProbEn~\cite{chen2022multimodal} or neighborhood‐saliency post‐processing~\cite{zhang2024cooperative}. Despite these advances, most methods are evaluated on only two modalities. SMART-Ship addresses this limitation by providing five rigorously registered modalities and dense, instance‐level polygon annotations, thereby enabling the design and fair comparison of more general and scalable fusion architectures.

\textbf{Cross‐Modal RS Re‐Identification Methods.}
Cross‐modal Re-ID in remote sensing seeks to match the same vessel instance across heterogeneous sensor domains. Existing algorithms map different modalities into a shared embedding space. SIDHCNN learns binary hash codes~\cite{li2018sidhcnn}, CMIR-NET employs cascaded attention for feature alignment~\cite{chaudhuri2020cmirnet}, DistillNet uses discriminative distillation to enhance inter‐class separability~\cite{xiong2020distill}, HGRMC-DS augments features via maximal correlation and Bayesian fusion~\cite{wang2024hgrmc}, and PRIML introduces proxy-based, rotation-invariant metric learning~\cite{cai2024}. While HOSS ReID extends instance-level labels to RGB–SAR pairs~\cite{gao2025hoss}, existing datasets lack both the modality diversity and fine‐grained annotations required for comprehensive evaluation. By supplying polygonal contours and unique instance identifiers across five modalities, SMART-Ship fills this gap and enables rigorous benchmarking of cross‐modal Re-ID approaches in maritime scenarios.

\textbf{Cross-Modal Generation Methods.} Cross-modal generation has evolved from conditional GANs (Pix2Pix~\cite{Pix2Pix2017}) to unpaired approaches using cycle-consistency (CycleGAN~\cite{CycleGAN2017}) or contrastive learning (CUT~\cite{CUT2020}). Recent advances incorporate semantic guidance (SelectionGAN~\cite{SelectionGAN2022}) or advanced network architectures (KAN-CUT~\cite{KANCUT2024}). For RS imagery, specialized methods address sensor-specific challenges through wavelet-based generation (WFLM-GAN~\cite{WflmGAN2022}) or semi-supervised frameworks (SemiD~\cite{SemiD2022}). However, existing methods primarily focus on single-modal pairs, lacking temporally synchronized multi-modal generation for maritime applications. Our SMART-Ship benchmark addresses this gap by providing registered multi-modal imagery with fine-grained annotations for rigorous evaluation under realistic maritime conditions.

\begin{figure*}
    \centering
    \includegraphics[width=0.95\linewidth]{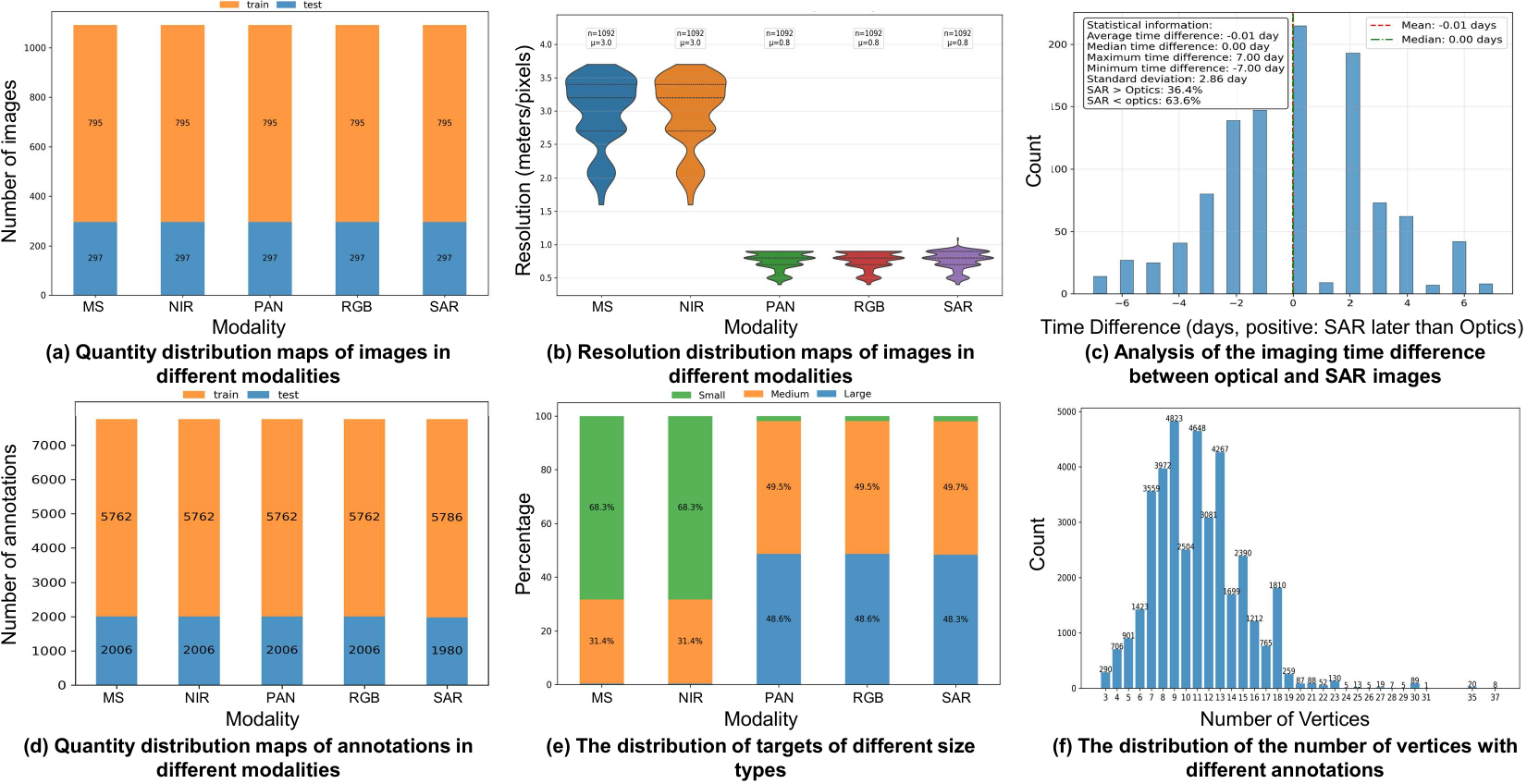}
    \caption{
    Overall Composition of the SMART-Ship Dataset.
    (a) image distribution across modalities;
    (b) ground sampling distance (GSD) distribution with low-resolution (MS, NIR: mean 3.0 m GSD) and high-resolution (PAN, RGB, SAR: mean 0.8m GSD) groups;
    (c) temporal acquisition analysis within a $\pm7$-day window;
    (d) distribution of polygon annotation counts across modalities;
    (e) target size distribution categorized by pixel dimensions (small: $\leq32 \times 32$, medium: $32 \times 32-96 \times 96$, large: $\geq 96 \times 96$ pixels);
    (f) vertex count distribution of polygon annotations, indicating fine-grained labeling quality with predominantly 9+ vertices per annotation.
}
    \label{fig_002}
\end{figure*}

\begin{figure*}
    \centering
    \includegraphics[width=0.95\linewidth]{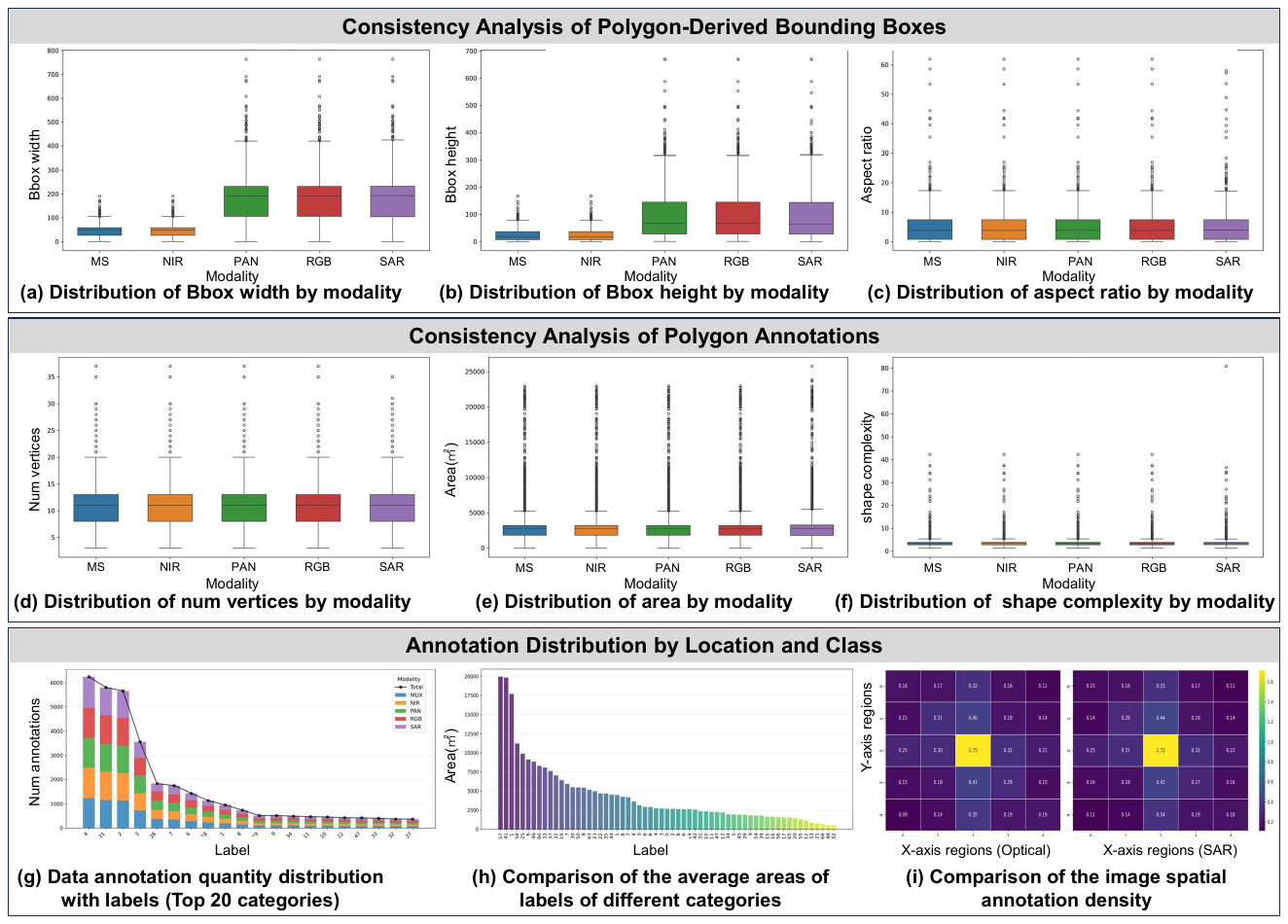}
    \caption{Multi-modal ship annotation statistics and consistency analysis.
    (a-c) Distribution of width, height, and aspect ratio of bounding rectangles from polygon annotations;
    (d) Distribution of vertex numbers in polygon annotations;
    (e) Area distribution of ground objects from polygon annotations;
    (f) Complexity distribution of polygon annotation shapes;
    (g) Annotation count distribution across ship categories (top 20);
    (h) Average annotated area distribution for different vessel types;
    (i) Spatial distribution heatmaps for optical and SAR image annotations, showing consistency between modalities.
    }
    \label{fig_003}
\end{figure*}

\textbf{Pan-Sharpening Methods.}
Pan-sharpening fuses low-resolution MS images with high-resolution PAN images to generate high-resolution RGB imagery, maintaining both spatial and spectral fidelity. The field has shifted from traditional statistical methods, such as IHS and PCA~\cite{pca}, and wavelet-based fusion, to deep learning-based approaches. Early deep learning models like PanNet~\cite{Yang_2017_ICCV} and DICNN~\cite{he2019dicnn} introduce residual learning and detail injection, improving high-frequency detail preservation. Subsequent networks, including BDPN~\cite{zhang2019pan} and MSDCNN~\cite{yuan2018multiscale}, leveraged multi-scale and hierarchical feature extraction for better fusion. Recent achievements, such as FusionNet~\cite{deng2020detail}, PNN~\cite{Ciotola2022}, and LAGNet~\cite{jin2022lagconv}, focus on adaptive optimization and context-aware fusion mechanisms. However, most pan-sharpening datasets mainly cover land and farmland scenes, lacking diversity in oceanic environments. To address this drawback, our SMART-Ship dataset offers extensive maritime and port imagery from multiple sensors. This enables comprehensive benchmarking of pan-sharpening methods in challenging marine conditions, where spectral mixing and atmospheric effects are pronounced.

\textbf{Cross-Modal Change Detection Methods.}
Existing cross-modal change detection methods can be broadly categorized into unsupervised, weakly supervised, and supervised approaches~\cite{dccd,eunet,subpixel}. Unsupervised methods typically attempt to align heterogeneous image representations in a common feature space without relying on labeled data, but often suffer from sensitivity to noise and false positives. Supervised methods utilize fully annotated datasets to train models for scene-level or instance-level change detection, but face challenges such as error propagation and inadequate handling of modality differences~\cite{dccd,eunet}. These issues can be partially attributed to the limited scale and quality of current datasets, which restrict the generalization and robustness of model training and evaluation. The SMART-Ship dataset addresses this gap by providing 1,092 pairs of registered, high-resolution RGB and SAR images with comprehensive instance-level change annotations. Based on these methods~\cite{ban,chen2021bit,cf,xie2021segformer}, we transferred a stackable architecture to evaluate the effectiveness of the SMART-Ship dataset.

\section{Dataset Description}
\label{sec:dataset}

\subsection{Overall description}
The proposed SMART-Ship dataset comprises five complementary modalities acquired from high-resolution satellite imagery, as illustrated in Figure~\ref{fig_001}. The dataset integrates MS, NIR, PAN, RGB, and SAR imagery, providing a comprehensive foundation for multi-modal remote sensing data analysis and maritime surveillance applications.

\textbf{Dataset Composition and Scale.} The dataset consists of 1,092 registered multi-modal image sets, with each set containing synchronized acquisitions across all five modalities. As shown in Figure~\ref{fig_002}, the dataset is divided into 795 training samples (72.8\%) and 297 testing samples (27.2\%), ensuring representation for both model development and evaluation. The temporal synchronization between modalities is maintained with an average time difference of -0.01 days and a standard deviation of 2.86 days, with all acquisitions constrained within a $\pm7$-day temporal window to ensure consistent environmental conditions. The dataset encompasses two distinct spatial resolution groups: low-resolution modalities (MS and NIR) with 3 meters average ground sampling distance, and high-resolution modalities (PAN, RGB, and SAR) with 0.8 meters average ground sampling distance, providing a resolution ratio of approximately $4:1$. This multi-resolution architecture enables comprehensive analysis of scale-dependent ship characteristics and supports multi-scale detection algorithms.

\subsection{Annotation Characteristics}
The dataset features high-precision polygonal annotations with an average of 12-13 vertices per ship, providing detailed geometric descriptions beyond conventional rectangular bounding boxes. Figure~\ref{fig_003} demonstrates comprehensive geometric analysis and consistency across all modalities, where Figure~\ref{fig_003} (a-c) compare the width, height, and aspect ratio distributions of bounding rectangles derived from polygonal annotations, while Figure~\ref{fig_003} (d-f) analyze the vertex count, polygon area (m$ ^2 $), and shape complexity ($\text{perimeter}^2 / \text{area}$) of the original polygonal annotations. 
Additionally, Figure~\ref{fig_003} (g-i) shows ship category distribution, vessel area statistics, and spatial density heatmaps, respectively. When comparing modalities, we primarily contrast optical imagery (RGB, PAN, MS) with SAR imagery due to their distinct sensing mechanisms.

\textbf{Annotation Geometric Analysis.} Geometric analysis (Figure~\ref{fig_003} (a-c)) reveals distinct scale characteristics between resolution groups. Low-resolution modalities (MS, NIR) exhibit median circumscribed rectangle dimensions of approximately 50×40 pixels, while high-resolution modalities (PAN, RGB, SAR) show median dimensions of 150-200×120-150 pixels, consistent with the 4:1 resolution ratio. The aspect ratio distribution remains consistent across all modalities with a median value of 1.5-2.0, indicating preserved geometric relationships despite resolution differences. 
Polygonal annotation analysis (Figure~\ref{fig_003} (d-e)) shows that ship annotations contain an average of 12-13 vertices per polygon, with polygon areas ranging from 1,000 to 10,000 $ \text{m}^2 $. The shape complexity (Figure~\ref{fig_003} (f)) analysis reveals that ship polygons exhibit high complexity values, reflecting the intricate geometric structures inherent in maritime vessel designs, and hull irregularities that distinguish ships from simple geometric shapes. These detailed geometric characteristics support precise ship localization and shape-aware detection algorithms.

\textbf{Annotation Spatial Distribution Analysis.} Since RGB, PAN, and MS are acquired from the same imaging platform, their annotations are perfectly aligned and can be collectively called optical imagery. Figure~\ref{fig_003} (i) demonstrates the spatial distribution density of annotation boxes across both optical and SAR modalities, revealing consistent spatial alignment characteristics between the two modalities. The observed concentration of peak density in the central regions is attributed to the dataset's target-centric cropping methodology, wherein 1024×1024 pixel images are extracted with targets positioned at the image center. This approach produces higher central annotation density with gradual peripheral decrease, ensuring balanced spatial coverage while maintaining target-focused characteristics.

\begin{table*}[t]
\centering
\captionsetup{justification=centering,singlelinecheck=false}
\caption{Task 1 Multi-Modal Ship Detection - quantitative performance of individual and fusion modalities.}
\label{tab:detection}
\resizebox{1\textwidth}{!}{
\begin{tabular}{l|c|c|c|c|c|c|ccc}
\toprule
\multirow{2}{*}{\textbf{Method}}
  & \multirow{2}{*}{\textbf{Domain}}
  & \multirow{2}{*}{\textbf{Year}}
  & \multirow{2}{*}{\textbf{Modality}}
  & \multirow{2}{*}{\textbf{Image Size}}
  & \multirow{2}{*}{\textbf{Params}}
  & \multirow{2}{*}{\textbf{Backbone}}
  & \multicolumn{3}{c}{\textbf{Detection Performance}} \\
\cmidrule(lr){8-10}
  & & & & & & & mAP@0.5$\uparrow$ & Precision$\uparrow$ & Recall$\uparrow$ \\
\midrule
\multicolumn{10}{c}{\cellcolor{gray!20}\textbf{(a) Single‑Modality Detection}} \\
\midrule
YOLOv5~\cite{jocher2021} & CV & 2020 & SAR  & $1,024\times1,024$ & 14.1M & CSPNet~\cite{wang2020cspnet} & 50.9 & \textbf{74.1} & 45.6 \\
YOLOv5~\cite{jocher2021} & CV & 2020 & RGB  & $1,024\times1,024$ & 14.1M & CSPNet~\cite{wang2020cspnet} & \textbf{68.7} & 67.0 & \textbf{62.8} \\
YOLOv5~\cite{jocher2021} & CV & 2020 & PAN  & $1,024\times1,024$ & 14.1M & CSPNet~\cite{wang2020cspnet} & \underline{59.7} & \underline{72.0} & 53.0 \\
YOLOv5~\cite{jocher2021} & CV & 2020 & MS   & $512\times512$   & 14.1M & CSPNet~\cite{wang2020cspnet} & 55.7 & 67.7 & 49.2 \\
YOLOv5~\cite{jocher2021} & CV & 2020 & NIR  & $512\times512$   & 14.1M & CSPNet~\cite{wang2020cspnet} & 58.5 & 65.7 & \underline{58.5} \\
\midrule
\multicolumn{10}{c}{\cellcolor{gray!20}\textbf{(b) RGB+SAR Fusion}} \\
\midrule
AddFusion~\cite{qingyun2022}    & RS & 2022 & RGB+SAR & $1,024\times1,024$ & 22.2M  & CSPNet~\cite{wang2020cspnet} & 56.9 & 59.0 & \underline{55.0} \\
CatFusion~\cite{qingyun2022}    & RS & 2022 & RGB+SAR & $1,024\times1,024$ & 23.5M  & CSPNet~\cite{wang2020cspnet} & 49.4 & 56.2 & 45.1 \\
InputFusion~\cite{qingyun2022}& RS & 2022 & RGB+SAR & $1,024\times1,024$ & 14.1M  & CSPNet~\cite{wang2020cspnet} & 51.8 & 63.6 & 48.2 \\
CFT~\cite{qingyun2021}          & RS & 2021 & RGB+SAR & $1,024\times1,024$ & 85.8M  & CFB~\cite{qingyun2021}     & \underline{60.3} & \underline{73.1} & 50.1 \\
ICAFusion~\cite{SHEN2024109913} & RS & 2024 & RGB+SAR & $1,024\times1,024$ & 230.9M & CSPNet~\cite{wang2020cspnet} & \textbf{66.1} & \textbf{82.5} & \textbf{57.5} \\
YOLOrs~\cite{sharma2021}        & RS & 2021 & RGB+SAR & $1,024\times1,024$ & 39.6M  & ResNet50~\cite{he2016deep}          & 56.4 & 84.7 & 47.8 \\
\midrule
\multicolumn{10}{c}{\cellcolor{gray!20}\textbf{(c) PAN+MS Fusion}} \\
\midrule
AddFusion~\cite{qingyun2022}    & RS & 2022 & PAN+MS & $1,024\times1,024$ & 22.2M  & CSPNet~\cite{wang2020cspnet} & 58.7 & 66.7 & 51.6 \\
CatFusion~\cite{qingyun2022}    & RS & 2022 & PAN+MS & $1,024\times1,024$ & 23.5M  & CSPNet~\cite{wang2020cspnet} & 57.9 & \underline{71.8} & 50.9 \\
InputFusion~\cite{qingyun2022}& RS & 2022 & PAN+MS & $1,024\times1,024$ & 14.1M  & CSPNet~\cite{wang2020cspnet} & 58.7 & 59.0 & 58.7 \\
CFT~\cite{qingyun2021}          & RS & 2021 & PAN+MS & $1,024\times1,024$ & 85.8M  & CFB~\cite{qingyun2021}     & \underline{62.6} & 59.6 & \underline{59.9} \\
ICAFusion~\cite{SHEN2024109913} & RS & 2024 & PAN+MS & $1,024\times1,024$ & 230.9M & CSPNet~\cite{wang2020cspnet} & \textbf{68.6} & \textbf{72.1} & \textbf{60.7} \\
YOLOrs~\cite{sharma2021}        & RS & 2021 & PAN+MS & $1,024\times1,024$ & 39.6M  & ResNet50~\cite{he2016deep}          & 56.6 & 86.8 & 47.0 \\
\bottomrule
\end{tabular}
}
\end{table*}

\textbf{Ship Category Distribution and Diversity.} The dataset encompasses 61 distinct ship categories with a total of 38,838 annotations across all modalities. Figure~\ref{fig_003} (g) illustrates the category distribution for the most frequent classes, revealing a pronounced long-tail distribution characteristic of SMART-Ship datasets. The distribution demonstrates significant class imbalance, with the top three categories (categories 4, 21, and 2) accounting for approximately 45.7\% of all annotations, while the remaining categories exhibit varying frequencies ranging from thousands to single-digit occurrences. 
This imbalanced distribution reflects the realistic ship type composition observed in port environments while supporting the development of class-aware detection methods and few-shot learning approaches.

\textbf{Ship Scale and Size Diversity Analysis.} 
Category-wise analysis of ship sizes, as presented in Figure~\ref{fig_003} (h), reveals substantial size diversity spanning from approximately 500 $ \text{m}^2 $ to 20,000 $ \text{m}^2 $ in real-world area. The largest category (label 57) exhibits an average area of approximately 20,000 $ \text{m}^2 $, while smaller categories approach 500 $ \text{m}^2 $, providing a size ratio of approximately 40:1. This extensive scale coverage enables comprehensive evaluation of multi-scale detection capabilities.

\textbf{Cross-Modal Consistency Analysis.} There is a high degree of consistency among different modalities. Temporal synchronization is verified with 63.6\% of SAR acquisitions preceding optical imagery and 36.4\% following, ensuring balanced temporal coverage (Figure~\ref{fig_002} (c)). Geometric consistency is validated through cross-modal annotation alignment, with identical polygon vertex counts and consistent aspect ratios across all modalities. The comprehensive coverage of ship sizes, categories, and spatial distributions provides a robust foundation for developing and evaluating multi-modal maritime ship detection algorithms. This multi-modal dataset addresses critical gaps in existing maritime RS datasets by providing comprehensive multi-modal coverage with five complementary modalities, high-precision polygonal annotations enabling detailed shape analysis, extensive scale and category diversity reflecting the ship composition in port environments, and rigorous temporal and spatial synchronization ensuring reliable multi-modal fusion capabilities.

\section{Benchmark Tasks and Evaluation}
\label{sec:benchmark}
We establish comprehensive benchmarks across five downstream tasks to evaluate the capabilities of the SMART-Ship dataset, each addressing different multi-modal analysis tasks.

\subsection{Task 1: Multi-Modal Ship Detection}
\label{subsec:detection}
Ship detection is a foundational task in RS, supporting critical applications such as maritime traffic monitoring, illegal activity prevention, resource protection, and search-and-rescue operations. However, this task is particularly challenging due to the inherent complexity of marine environments. To address these challenges, recent advances have explored the use of multi-modal RS data, each modality provides distinct advantages. By fusing complementary information across these modalities, it becomes possible to improve detection robustness under diverse and adverse acquisition conditions.

Despite these opportunities, existing object detection frameworks often degrade when directly applied to multi-modal satellite imagery, primarily due to the heterogeneity of sensor characteristics and the insufficient exploration of different modal fusions. In this study, we present an evaluation of fusion-based object detectors based on the proposed SMART-Ship dataset.

\subsubsection{Task Formulation} 
Formally, given a set of registered images $\mathbf{x} = \{x^{(m)}\}_{m=1}^M$ from $M$ distinct modalities, where each image $x^{(m)} \in \mathcal{X}^{(m)}$ belongs to the domain of modality $m$. The objective is to learn a detection function $f_\theta(\cdot)$ parameterized by $\theta$. This function predicts a set of bounding boxes $\mathcal{B} = \{b_i\}_{i=1}^K$ with associated class labels $\mathcal{C} = \{c_i\}_{i=1}^K$ and confidence scores $\mathcal{S} = \{s_i\}_{i=1}^K$, where $K$ denotes the number of detected ships. Each $b_i$ is defined by the coordinates $(x_i, y_i, w_i, h_i)$ representing the location and size of the object. Mathematically:
\begin{equation}
    f_\theta : \mathbf{x} \rightarrow \{(b_i, c_i, s_i)\}_{i=1}^K
\end{equation}

\begin{figure*}[htbp]
    \centering
    \includegraphics[width=0.95\linewidth]{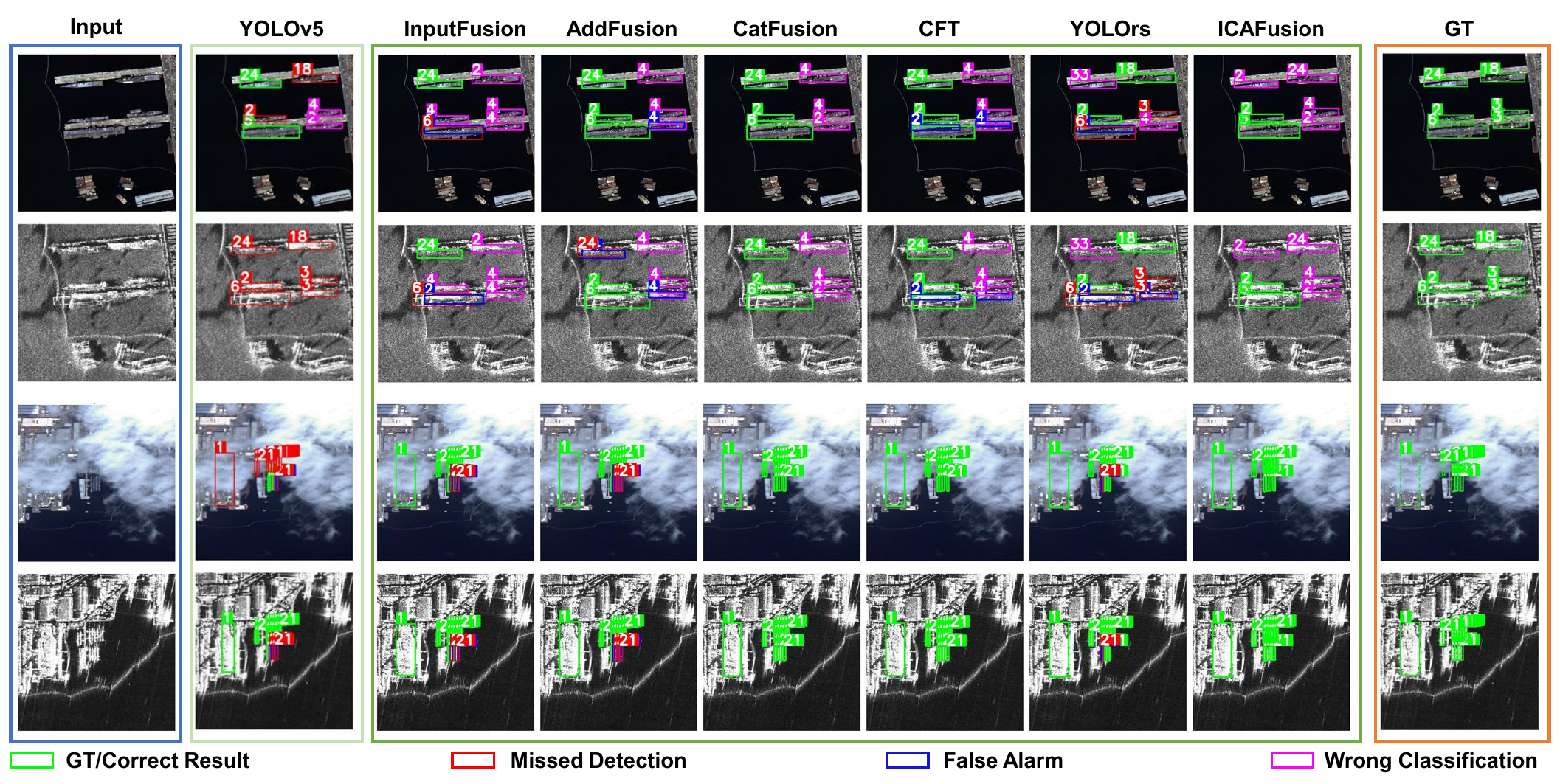}
    \caption{
    Task 1 Multi-Modal Ship Detection - qualitative comparison on RGB and SAR modalities.
    The first column shows the input data; the second column presents single‐modality detection results; columns three through seven illustrate the outputs of six distinct fusion‐based detection approaches, each compared against the GT annotations.}
    \label{fig_004}
\end{figure*}

\subsubsection{Evaluation Setting}
\textbf{Dataset.} 
Our experiments are conducted on the proposed SMART-Ship dataset. The dataset is split by image IDs into training (00001--00795) and testing (00796--01092) sets. For single-modality detection, models are trained and evaluated independently on each modality following classical object detection benchmarks~\cite{jocher2021}. For the fusion detection experiments, we focus on two representative modality pairs, i.e., RGB+SAR and PAN+MS, and each sample comprises the spatially registered images. Because the MS images are originally acquired at a lower spatial resolution, they are resampled with interpolation to match the ground sampling distance of the PAN imagery, thereby avoiding scale‑induced bias during training and evaluation.

\textbf{Implementation Details.} 
Owing to the spatiotemporal consistency of the RGB, PAN, MS, and NIR images, the captured images from these sensors share the same label. Therefore, the labels are equal to single-modality for the PAN+MS fusion settings. For the RGB+SAR fusion experiments, temporal misalignment induces inconsistencies across modalities, potentially resulting in object displacement, disappearance, or the emergence of new objects. Hence, we match the objects between optical and SAR images based on class labels and Intersection over Union (IoU) of HBBs. We define the fused target as the minimum enclosing rectangle of the matched objects, ensuring that the final label fully covers the object in both modalities. Both fused targets and single‑modality‑specific targets are included in the final ground‑truth set. This inclusive annotation strategy encourages detectors to maximize recall under modality‑specific visibility gaps while still rewarding precise localization. 

We benchmark multiple methods, including the popular single‐modal detector YOLOv5~\cite{jocher2021}, in order to compare their performance directly against our fusion‐based detection approaches and quantify the benefits of multi‐modal data fusion. Fusion-based methods includes InputFusion~\cite{qingyun2022}, AddFusion~\cite{qingyun2022}, CatFusion~\cite{qingyun2022}, Cross-Modality Fusion Transformer (CFT)~\cite{qingyun2021}, YOLOrs~\cite{sharma2021}, and ICAFusion~\cite{SHEN2024109913}. 
For ICAFusion, YOLOrs, CFT, and YOLOv5, we strictly adhere to the original hyperparameter settings. For InputFusion, AddFusion, and CatFusion, due to lack of original hyper-parameters, we adopt those reported in~\cite{zhangCADDN}, including: SGD optimizer momentum 0.98; warm-up for the first three epochs; cosine annealing learning rate schedule from 0.01 to 0.001; loss weights for regression, confidence, and classification losses set to 0.05, 2.0, and 0.5 respectively; prediction layer weights as 4.0, 1.0, and 0.4; positive sample weight 2.0; and weight decay 0.0005. 

\textbf{Evaluation Metrics.}
Following the evaluation protocols of existing RS image object detection works~\cite{qingyun2021,qingyun2022,zhangCADDN, SODA, VisDrone, ding2021object, yang2022scrdet++,xiao2024highly,zhao2025hybrid}, we employ precision (P), recall (R), and mean average precision at an IoU threshold of 0.5 (mAP@0.5) to evaluate detection performance quantitatively.

\subsubsection{Experimental Results}
\textbf{Single‐Modality Detection Performance.}  
As shown in Table~\ref{tab:detection} (a), the RGB detector achieves the highest mAP@0.5 and recall on the test set, whereas the SAR detector exhibits the lowest performance. This disparity can be attributed to the rich color, as well as the higher spatial resolution of the RGB imagery, which facilitates feature discrimination and localization. In contrast, SAR images lack fine-grained texture and color information, leading to inferior overall accuracy under clear sky conditions. Nonetheless, in occluded scenes, e.g., the example in the bottom two rows of Figure~\ref{fig_004}, the all‐weather imaging capability of SAR proves advantageous, allowing it to detect ships where RGB performance degrades.

\begin{table*}[t]
\centering
\caption{Task 2 Cross-Modal Ship Re-Identification - quantitative comparison across different modalities.}
\label{tab:crossmodal_reid}
\resizebox{1\textwidth}{!}{
\begin{tabular}{l|c|c|c|ccc|ccc|ccc}
\toprule
\multirow{2}{*}{\textbf{Method}} 
  & \multirow{2}{*}{\textbf{Domain}} 
  & \multirow{2}{*}{\textbf{Year}} 
  & \multirow{2}{*}{\textbf{Params}} 
  & \multicolumn{3}{c|}{\textbf{Forward (A $\rightarrow$ B)}} 
  & \multicolumn{3}{c|}{\textbf{Backward (B $\rightarrow$ A)}}
  & \multicolumn{3}{c}{\textbf{Mean}} \\
\cmidrule(lr){5-7}\cmidrule(lr){8-10}\cmidrule(lr){11-13}
  & & & & mAP$\uparrow$ & Rank1$\uparrow$ & Rank10$\uparrow$ 
  & mAP$\uparrow$ & Rank1$\uparrow$ & Rank10$\uparrow$ 
  & mAP$\uparrow$ & Rank1$\uparrow$ & Rank10$\uparrow$ \\
\midrule
\rowcolor{gray!20}
\multicolumn{13}{c}{\textbf{(a) RGB $\leftrightarrow$ SAR}} \\
\midrule
Unified loss~\cite{sun2020}      & CV & 2020 & 98.4M & 36.2                & \textbf{42.8}          & 83.7                & 37.1                & 39.7                   & 80.3           & 36.7 & \textbf{41.3} & 82.0 \\
Circle loss~\cite{sun2020}       & CV & 2020 & 98.4M & 37.3                & 39.2                   & \underline{84.7}    & 37.4                & 37.0                   & 80.5           & 37.4 & 38.1 & \underline{82.6} \\
Identity loss~\cite{zheng2017}   & CV & 2017 & 98.4M & \textbf{37.9}       & 40.1                   & 82.7                & \textbf{39.4}       & \textbf{41.6}          & \underline{81.0}  & \textbf{38.7} & \underline{40.9} & 81.9 \\
Sphere loss~\cite{liu2017}       & CV & 2017 & 98.4M & 36.7                & 38.4                   & 83.5                & \underline{38.4}    & \underline{39.9}       & 80.3           & \underline{37.6} & 39.2 & 81.9 \\
Lifted loss~\cite{song2016}      & CV & 2016 & 98.4M & 36.4                & 38.2                   & 82.2                & 37.8                & 39.2                   & 80.3           & 37.1 & 38.7 & 81.3 \\
Instance loss~\cite{zheng2020}   & CV & 2020 & 98.4M & 36.8                & 38.9                   & 81.8                & 37.3                & 38.4                   & \underline{81.0} & 37.1 & 38.7 & 81.4 \\
PBFFN~\cite{cai2024}             & RS & 2024 & 98.4M & \underline{35.8}    & \underline{39.9}       & \textbf{85.4}       & 38.3                & 37.7                   & \textbf{83.0}  & 37.1 & 38.8 & \textbf{84.2} \\
\midrule
\rowcolor{gray!20}
\multicolumn{13}{c}{\textbf{(b) RGB $\leftrightarrow$ PAN}} \\
\midrule
Unified loss~\cite{sun2020}      & CV & 2020 & 98.4M & 80.2                & 94.4                   & \underline{99.3}    & 80.9                & 95.6                   & \underline{99.5} & 80.6 & 95.0 & 99.4 \\
Circle loss~\cite{sun2020}       & CV & 2020 & 98.4M & 80.7                & \underline{96.1}       & \underline{99.3}    & 81.0                & \underline{96.6}       & 99.3           & 80.9 & \textbf{96.4} & 99.3 \\
Identity loss~\cite{zheng2017}   & CV & 2017 & 98.4M & \underline{81.5}    & \textbf{96.4}          & \underline{99.3}    & 81.2                & 96.1                   & \underline{99.5} & 81.4 & \underline{96.3} & 99.4 \\
Sphere loss~\cite{liu2017}       & CV & 2017 & 98.4M & \textbf{82.3}       & 95.9                   & 99.0                & \underline{81.9}    & 96.4                   & 99.3           & \underline{82.1} & 96.2 & 99.2 \\
Lifted loss~\cite{song2016}      & CV & 2016 & 98.4M & 81.4                & 95.9                   & \textbf{99.5}       & 81.3                & 96.4                   & \underline{99.5} & 81.4 & 96.2 & 99.5 \\
Instance loss~\cite{zheng2020}   & CV & 2020 & 98.4M & \textbf{82.3}       & 95.1                   & \underline{99.3}    & \textbf{82.1}       & \textbf{96.9}          & \textbf{99.8}  & \textbf{82.2} & 96.0 & \underline{99.6} \\
PBFFN~\cite{cai2024}             & RS & 2024 & 98.4M & 80.7                & 95.6                   & \textbf{99.5}       & 80.8                & 95.9                   & \textbf{99.8}  & 80.8 & 95.8 & \textbf{99.7} \\
\midrule
\rowcolor{gray!20}
\multicolumn{13}{c}{\textbf{(c) SAR $\leftrightarrow$ PAN}} \\
\midrule
Unified loss~\cite{sun2020}      & CV & 2020 & 98.4M & \textbf{37.8}       & \textbf{38.2}          & 73.7                & \textbf{36.6}       & \underline{38.9}       & 82.0           & \textbf{37.2} & \textbf{38.6} & 77.9 \\
Circle loss~\cite{sun2020}       & CV & 2020 & 98.4M & 35.3                & 35.3                   & \underline{77.1}    & 35.7                & 38.0                   & \underline{82.7} & 35.5 & 36.7 & \underline{79.9} \\
Identity loss~\cite{zheng2017}   & CV & 2017 & 98.4M & 35.5                & 34.5                   & 74.7                & 35.0                & 38.4                   & 80.0           & 35.3 & 36.5 & 77.4 \\
Sphere loss~\cite{liu2017}       & CV & 2017 & 98.4M & 37.3                & 36.7                   & 75.9                & \underline{35.9}    & 38.7                   & \textbf{83.5}  & 36.6 & \underline{37.7} & 79.7 \\
Lifted loss~\cite{song2016}      & CV & 2016 & 98.4M & \underline{37.7}    & \underline{38.0}       & \textbf{79.6}       & 35.7                & 36.3                   & 82.2           & \underline{36.7} & 37.2 & \textbf{80.9} \\
Instance loss~\cite{zheng2020}   & CV & 2020 & 98.4M & 36.6                & 36.0                   & 74.2                & 35.3                & 37.2                   & 82.5           & 36.0 & 36.6 & 78.4 \\
PBFFN~\cite{cai2024}             & RS & 2024 & 98.4M & 34.5                & 33.1                   & 75.2                & 35.7                & \textbf{41.1}          & 79.6           & 35.1 & 37.1 & 77.4 \\
\bottomrule
\end{tabular}
}

\vspace{5pt}
\begin{minipage}{\textwidth}
\footnotesize
Note: The scarcity of RS-specific methods reflects limited open-source implementations and benchmark datasets for the Re-ID task.
\end{minipage}
\end{table*}

\textbf{Fusion Method Performance Analysis.}  
Table~\ref{tab:detection}(b)–(c) reveals substantial variation among fusion strategies for each modality pair. In both RGB+SAR and PAN+MS settings, ICAFusion~\cite{SHEN2024109913} achieves the highest performance, underscoring the critical importance of advanced feature‐alignment and integration modules. This result demonstrates that carefully designed fusion architectures are essential for effective multi‐modal ship detection and suggests that there remains significant room to improve. Moreover, we observe that RGB+SAR fusion performance is lower than PAN+MS. This gap likely results from the larger spectral domain discrepancy and temporal misalignment between RGB and SAR acquisitions, which can produce inconsistent target appearances and hinder detector performance. Therefore, we advocate for future research to develop more robust RGB+SAR detection methods.

\textbf{Fusion vs.\ Single‐Modality Comparison.}  
When comparing fusion against single‐modality baselines, we observe that RGB+SAR fusion does not yield a clear overall advantage over the RGB‐only detector, suggesting that the simple combination of highly heterogeneous modalities can introduce noise unless mitigated by targeted alignment. Conversely, PAN+MS fusion schemes deliver significant gains relative to their single‐modality baseline YOLOv5~\cite{jocher2021}, highlighting the benefit of combining spectrally and spatially complementary bands. Notably, certain challenging scenes benefit from RGB+SAR fusion, e.g., scenarios in Figure~\ref{fig_004}, where the fusion detector leverages both the detailed optical appearance and the penetration capability of SAR. This observation indicates that well‑designed fusion strategies can effectively exploit the complementary strengths of optical and SAR modalities to improve detection performance.

Overall, our benchmark validates the critical role of multi‐modal fusion in maritime ship detection and emphasizes that fusion effectiveness strongly depends on modality affinity and the sophistication of the fusion architecture.

\subsection{Task 2: Cross-Modal Ship Re-Identification}
\label{subsec:reid}
Ship Re-ID aims to match observations of the same ship captured under different modalities. This capability is of fundamental importance in real-world maritime surveillance systems, where persistent ship monitoring must operate across diverse environments based on complementary sensors. Furthermore, effective cross-modal ship Re-ID enables several downstream applications, including long-term trajectory reconstruction, multi-sensor behavior analysis, and multi-view fusion for maritime situational awareness. These capabilities are increasingly relevant in the era of growing satellite coverage and sensor heterogeneity.

Unlike conventional single-modality Re-ID, cross-modal ship Re-ID focuses on retrieving ship identities across heterogeneous domains, bridging significant visual and statistical gaps between modalities. The cross-modal setting poses unique challenges due to substantial domain shifts in appearance, geometry, and sensing noise. However, to the best of our knowledge, there is no benchmark for five-modalities RS ship Re-ID due to data limitations. To this end, we evaluate representative metric learning-based Re-ID methods under three cross-modal scenarios of practical significance: RGB$\leftrightarrow$SAR, PAN$\leftrightarrow$RGB, and SAR$\leftrightarrow$PAN.

\begin{figure*}[htbp]
    \centering
    \includegraphics[width=0.95\linewidth]{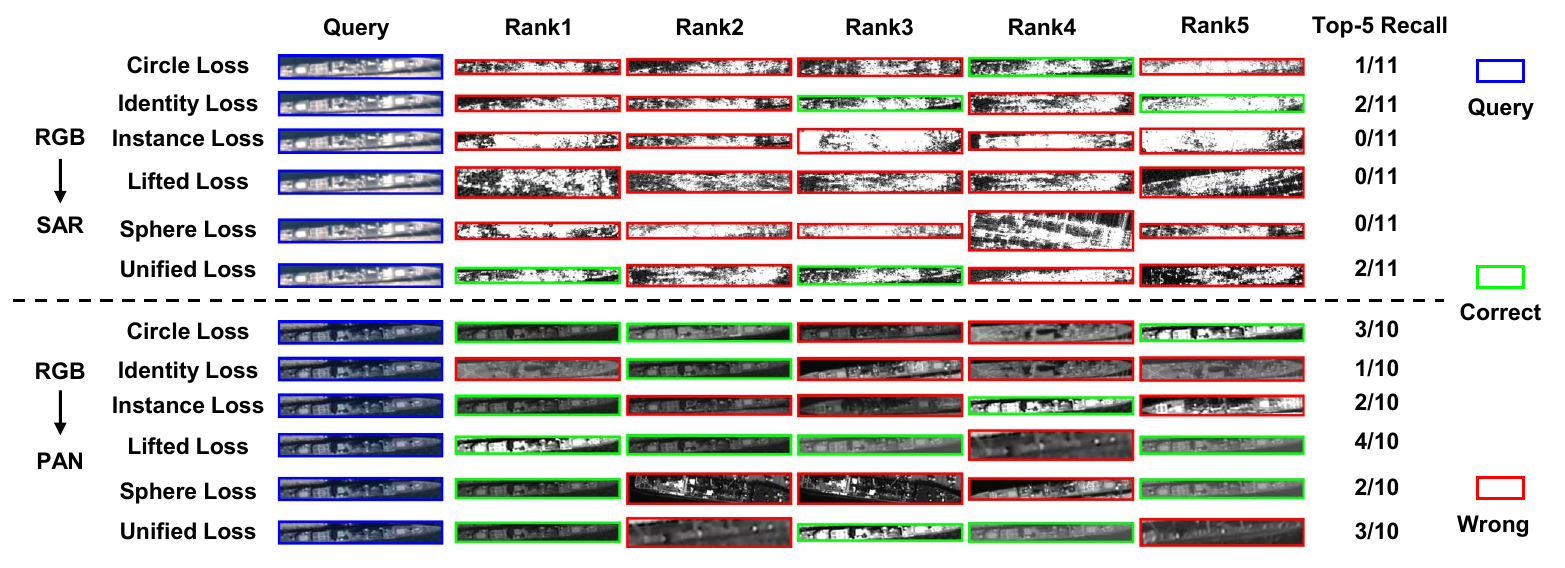}
    \caption{Task 2: Cross-Modal Ship Re-Identification - qualitative results from RGB to SAR, and RGB to PAN. Each row displays the top-5 retrieval results from the gallery using different representative methods.
    }
    \label{fig_005}
\end{figure*}

\subsubsection{Task Formulation}
\label{subsec:reid_form}
Given two modalities ${m}_{a}$ and ${m}_{b}$, a training set 
$\mathcal{T}=\{(x_i,y_i)\}_{i=1}^{N}$ provides image $x_i$ and its identity label  
$y_i\in\{1,\dots,C\}$.  At test time the query set $\mathcal{Q}^{{m}_{a}}=\{q_j\}_{j=1}^{|\mathcal{Q}|}$ from modality ${m}_{a}$ and the gallery set
$\mathcal{G}^{{m}_{b}}=\{g_k\}_{k=1}^{|\mathcal{G}|}$ from modality ${m}_{b}$ are provided, where $\mathcal{Q}\cap\mathcal{G}=\varnothing$.  
A shared CNN encoder converts an image into a $d$‑dimensional embedding. In the standard Re-ID workflow, feature embeddings of each query and all gallery images are compared using cosine similarity, and the resulting scores are sorted in descending order to produce the final ranking.

\subsubsection{Evaluation Setting}
\label{subsec:reid_setting}
\textbf{Dataset.}  
To rigorously assess cross-modal generalization, we adopt a disjoint identity split strategy. Specifically, for each class indexed from 1 to 50, we randomly select 70\% of its images for the training set and allocate the remaining 30\% to intra-class testing. All images from classes with indices greater than 50 are excluded from the training set and reserved exclusively for testing. This identity separation provides a zero-shot retrieval scenario and evaluates the generalization capability. We focus on three practically significant cross-modal retrieval directions in maritime settings: RGB$\leftrightarrow$SAR, PAN$\leftrightarrow$RGB, and SAR$\leftrightarrow$PAN.

\textbf{Implementation Details.}  
All images are resized to $256\times256$ pixels to normalize the diverse object scales encountered in remote sensing and to match the fixed input dimensions required by our neural network. We mainly evaluate popular metric losses from computer vision, including {Circle loss}~\cite{sun2020}, {Unified loss}~\cite{sun2020}, {Identity loss}~\cite{zheng2017}, {Sphere loss}~\cite{liu2017}, {Lifted loss}~\cite{song2016},  and {Instance loss}~\cite{zheng2020}. Considering the scarcity of cross-modal ReID works in remote sensing, we selected the state-of-the-art cross-modal image retrieval method PBFFN~\cite{cai2024} for experimentation. We adopt ResNet‑50 with an embedding dimension $d{=}64$. SGD is used with momentum $0.9$ and weight decay $0.0005$. Training proceeds in 100 epochs with learning rates $0.05$ and $0.005$. The batch size is $64$ and the temperature coefficient for metric losses is $\nu{=}0.001$.  

\textbf{Evaluation Metrics.}  
Following the experimental protocol of standard Re‐ID studies~\cite{meng2021deep, Zhou_2023_CVPR,ye2021deep, CTAL, UTAL, ye2023channel}, we employ mean Average Precision (mAP), Rank-1 accuracy (Rank1), and Rank-10 accuracy (Rank10) to evaluate the performance of each method.

\subsubsection{Experimental Results}
\label{subsec:reid_analysis}

\textbf{Performance Across Modality Pairs.}  
Table~\ref{tab:crossmodal_reid} summarizes the retrieval accuracy for three cross-modal scenarios. The RGB\,$\leftrightarrow$\, SAR and SAR\,$\leftrightarrow$\,PAN tasks exhibit low performance, reflecting the substantial appearance and statistical gap between RGB and SAR images. PAN\,$\leftrightarrow$\, RGB achieves the highest retrieval performance, due to the closer spectral‐spatial characteristics of PAN and RGB sensors. These results confirm that cross‐modal Re-ID performance strongly correlates with the inherent affinity of the sensor pair.

\textbf{Different Methods Comparison.}  
Although the evaluated metric‐learning losses exhibit varying strengths on different sensor pairs, our comprehensive benchmark finds that no single method consistently achieves SOTA performance across all cross‐modal scenarios. Instead, the choice of modality pair accounts for the majority of performance variation, indicating that modality affinity exerts a more profound influence on retrieval accuracy than the specific loss‐function design. Since most existing Re-ID approaches are originally developed for pedestrian retrieval, there remains a clear need for novel architectures and loss formulations tailored to the unique challenges of heterogeneous remote sensing cross-modal Re-ID, particularly in challenging situations such as RGB$\leftrightarrow$SAR and PAN$\leftrightarrow$SAR.

\textbf{Forward vs.\ Backward Retrieval Symmetry.}  
Across all methods and modality pairs, forward (A\,$\rightarrow$\,B) and backward (B\,$\rightarrow$\,A) retrieval show minor performance asymmetries (typically 1–2\% in mAP). This near‐symmetry indicates that the shared embedding space learned by the CNN encoder generalizes consistently in both directions, although slight imbalance arises when one modality presents more challenging appearance variations (e.g., RGB imagery tends to offer richer texture cues than SAR, improving SAR\,$\rightarrow$\,RGB performance marginally over the reverse).

These observations imply that future work should prioritize specialized cross‐domain alignment modules for highly heterogeneous sensor pairs. Our proposed benchmark thus provides a rigorous foundation for future research in maritime ship Re-ID, guiding the design of next‐generation algorithms tailored to heterogeneous satellite data.

\subsection{Task 3: Cross-Modal Generation}
\label{subsec:generation}

\begin{table*}[t]
\centering
\caption{Task 3 Cross-Modal Generation - quantitative comparison of different methods across different modalities.}
\label{tab:multimodal_comparison}
\begin{tabular}{l|cc|c|ccc|ccc|ccc}
\toprule
\multirow{2}{*}{\textbf{Method}} & \multirow{2}{*}{\textbf{Domain}} & \multirow{2}{*}{\textbf{Year}} & \multirow{2}{*}{\textbf{Params}} & \multicolumn{3}{c|}{\textbf{Forward (A $\rightarrow$ B)}} & \multicolumn{3}{c|}{\textbf{Backward (B $\rightarrow$ A)}} & \multicolumn{3}{c}{\textbf{Mean}} \\
\cmidrule(lr){5-7} \cmidrule(lr){8-10} \cmidrule(lr){11-13}
 & & & & PSNR$\uparrow$ & SSIM$\uparrow$ & FID$\downarrow$ & PSNR$\uparrow$ & SSIM$\uparrow$ & FID$\downarrow$ & PSNR$\uparrow$ & SSIM$\uparrow$ & FID$\downarrow$ \\
\midrule
\multicolumn{13}{c}{\cellcolor{gray!20}\textbf{(a) RGB $\leftrightarrow$ SAR}} \\
\midrule
Pix2Pix~\cite{Pix2Pix2017} & CV & 2017 & 54.412M & 12.65 & 0.155 & 166.50 & 12.65 & 0.308 & \underline{264.15} & 12.65 & 0.231 & 215.33 \\
CycleGAN~\cite{CycleGAN2017} & CV & 2017 & 22.744M & 10.86 & 0.111 & \textbf{108.47} & 11.24 & 0.227 & \textbf{211.87} & 11.05 & 0.169 & \textbf{160.17} \\
CUT~\cite{CUT2020} & CV & 2020 & 14.137M & 10.78 & 0.107 & 123.72 & 9.94 & 0.183 & 288.63 & 10.36 & 0.145 & 206.18 \\
SelectionGAN~\cite{SelectionGAN2022} & CV & 2022 & 58.244M & 13.78 & 0.193 & 361.88 & \textbf{13.61} & 0.346 & 427.31 & \underline{13.69} & 0.270 & 394.59 \\
KANCUT~\cite{KANCUT2024} & CV & 2024 & 14.129M & 11.23 & 0.103 & \underline{119.56} & 11.09 & 0.242 & 268.78 & 11.16 & 0.172 & \underline{194.17} \\
\midrule
WflmGAN~\cite{WflmGAN2022} & RS & 2022 & 221.231M & \textbf{14.24} & \underline{0.237} & 362.78 & \underline{13.51} & \textbf{0.433} & 426.33 & \textbf{13.88} & \textbf{0.335} & 394.55 \\
SemiD2~\cite{SemiD2022} & RS & 2022 & 33.800M & \underline{13.95} & \textbf{0.238} & 333.92 & 13.05 & \underline{0.391} & 428.16 & 13.50 & \underline{0.315} & 381.04 \\
\midrule
\multicolumn{13}{c}{\cellcolor{gray!20}\textbf{(b) RGB $\leftrightarrow$ PAN}} \\
\midrule
Pix2Pix~\cite{Pix2Pix2017} & CV & 2017 & 54.412M & 12.25 & 0.346 & 310.55 & 13.53 & 0.337 & \underline{289.93} & 12.89 & 0.341 & 300.24 \\
CycleGAN~\cite{CycleGAN2017} & CV & 2017 & 22.744M & 11.35 & 0.323 & \textbf{59.234} & 12.26 & 0.301 & \textbf{87.503} & 11.81 & 0.312 & \textbf{73.37} \\
CUT~\cite{CUT2020} & CV & 2020 & 14.137M & 11.32 & 0.316 & \underline{150.04} & 10.97 & 0.224 & 292.41 & 11.14 & 0.270 & \underline{221.23} \\
SelectionGAN~\cite{SelectionGAN2022} & CV & 2022 & 58.244M  & \textbf{13.95} & \underline{0.423} & 475.93 & 14.62 & 0.374 & 429.06 & \textbf{14.29} & 0.399 & 452.50 \\
KANCUT~\cite{KANCUT2024} & CV & 2024 & 14.129M & 11.07 & 0.303 & 150.82 & 11.00 & 0.220 & 323.08 & 11.04 & 0.261 & 236.95 \\
\midrule
WflmGAN~\cite{WflmGAN2022} & RS & 2022 & 221.231M & \underline{13.26} & \textbf{0.447} & 337.12 & \textbf{15.03} & \textbf{0.464} & 367.21 & \underline{14.14} & \textbf{0.456} & 352.16 \\
SemiD2~\cite{SemiD2022} & RS & 2022 & 33.800M & 13.22 & 0.420 & 333.81 & \underline{14.63} & \underline{0.438} & 354.53 & 13.93 & \underline{0.429} & 344.17 \\
\midrule
\multicolumn{13}{c}{\cellcolor{gray!20}\textbf{(c) SAR $\leftrightarrow$ PAN}} \\
\midrule
Pix2Pix~\cite{Pix2Pix2017} & CV & 2017 & 54.412M & 12.09 & 0.325 & 251.31 & 12.35 & 0.146 & 153.94 & 12.22 & 0.236 & 202.63 \\
CycleGAN~\cite{CycleGAN2017} & CV & 2017 & 22.744M & 10.91 & 0.284 & 162.36 & 10.91 & 0.110 & \textbf{101.71} & 10.91 & 0.197 & \textbf{132.04} \\
CUT~\cite{CUT2020} & CV & 2020 & 14.137M & 10.76 & 0.294 & \textbf{152.36} & 11.24 & 0.115 & \underline{126.47} & 11.00 & 0.205 & \underline{139.41} \\
SelectionGAN~\cite{SelectionGAN2022} & CV & 2022 & 58.244M  & \textbf{13.09} & 0.389 & 447.83 & 13.53 & \underline{0.190} & 371.91 & \textbf{13.31} & 0.289 & 409.87 \\
KANCUT~\cite{KANCUT2024} & CV & 2024 & 14.129M& 10.86 & 0.297 & \underline{159.70} & 11.24 & 0.109 & 128.74 & 11.05 & 0.203 & 144.22 \\
\midrule
WflmGAN~\cite{WflmGAN2022} & RS & 2022 & 221.231M & \underline{12.70} & \textbf{0.424} & 334.60 & \textbf{13.89} & \textbf{0.234} & 366.32 & \underline{13.30} & \textbf{0.329} & 350.46 \\
SemiD2~\cite{SemiD2022} & RS & 2022 & 33.800M & 12.57 & \underline{0.395} & 400.36 & \underline{13.83} & \textbf{0.234} & 352.45 & 13.20 & \underline{0.314} & 376.41 \\
\bottomrule
\end{tabular}
\end{table*}

Cross-modal generation facilitates the synthesis of one modality from another, offering significant advantages for data augmentation, missing modality compensation, and multi-modal analysis in RS applications. This study evaluates three bidirectional generation tasks: (1) RGB$\leftrightarrow$SAR generation enables all-weather earth observation by bridging optical-radar domains for continuous monitoring. 
(2) RGB$\leftrightarrow$PAN generation enhances spatial-spectral fusion by integrating high-resolution PAN imagery with MS information. 
(3) SAR$\leftrightarrow$PAN generation combines complementary sensing mechanisms, merging the surface penetration capabilities of radar with PAN spatial details.

\subsubsection{Task Formulation}
Given an input image $x^{(m_s)}$ from source modality $m_s \in \mathcal{M}$, cross-modal generation synthesizes a corresponding image $\hat{y}^{(m_t)}$ in target modality $m_t \in \mathcal{M}$ preserving structural and semantic content. This learns an optimal mapping $G: \mathcal{X}^{(m_s)} \rightarrow \mathcal{X}^{(m_t)}$ where $\mathcal{X}^{(m)}$ denotes the image domain of modality $m$:
\begin{equation}
    \hat{y}^{(m_t)} = G(x^{(m_s)}), \quad x^{(m_s)} \in \mathcal{X}^{(m_s)}, \hat{y}^{(m_t)} \in \mathcal{X}^{(m_t)}
\end{equation}
The model training process employs paired samples $\mathcal{T} = \{(x_{t,i}^{(m_s)}, y_{t,i}^{(m_t)})\}_{i=1}^N$ with registered $\{x_{t,i}^{(m_s)}, y_{t,i}^{(m_t)}\}$ from identical geographical locations.

\subsubsection{Evaluation Setting}
\textbf{Dataset.} Experiments utilize a multi-modal RS dataset with 1,092 registered image triplets (RGB, SAR, PAN), following the overall dataset partitioning scheme: 795 training and 297 testing samples. All images are standardized to a resolution of 256×256 pixels to ensure consistent processing across modalities.

\textbf{Implementation Details.}
We implement and evaluate seven representative cross-modal generation methods spanning different technical approaches: (1) CycleGAN and Pix2Pix as classical image-to-image generation frameworks; (2) CUT as a contrastive learning-based approach; (3) SemiD2 and WFLMGAN as RS specialized methods; (4) SelectionGAN as a selection-based method; (5) KAN-CUT as a recent kernel-based approach.

We conduct comprehensive experiments for each model across three bidirectional modality generation tasks, resulting in six distinct generation directions. To evaluate parameter sensitivity and optimization stability, each model configuration is trained with two different learning rates (0.0002 and 0.0005), maintaining consistent hyperparameters.
All models are trained for 200 epochs using the Adam optimizer with $\beta_1=0.5$ and $\beta_2=0.999$. A linear learning rate decay policy is applied, starting from the initial learning rate and gradually decreasing to zero over the training period. The batch size is set to 1 across all experiments due to memory constraints and to maintain consistency with the original implementations of the baseline methods. For data augmentation, we employ random cropping and horizontal flipping during training.

\textbf{Evaluation Metrics.}
We evaluate generated images using three complementary metrics: Peak Signal-to-Noise Ratio (PSNR) for pixel-level fidelity, Structural Similarity Index (SSIM) for structural preservation, and Fréchet Inception Distance (FID~\cite{heusel2017gans}) for perceptual quality.

\subsubsection{Experimental Results}
\begin{figure*}
    \centering
    \includegraphics[width=0.9\linewidth]{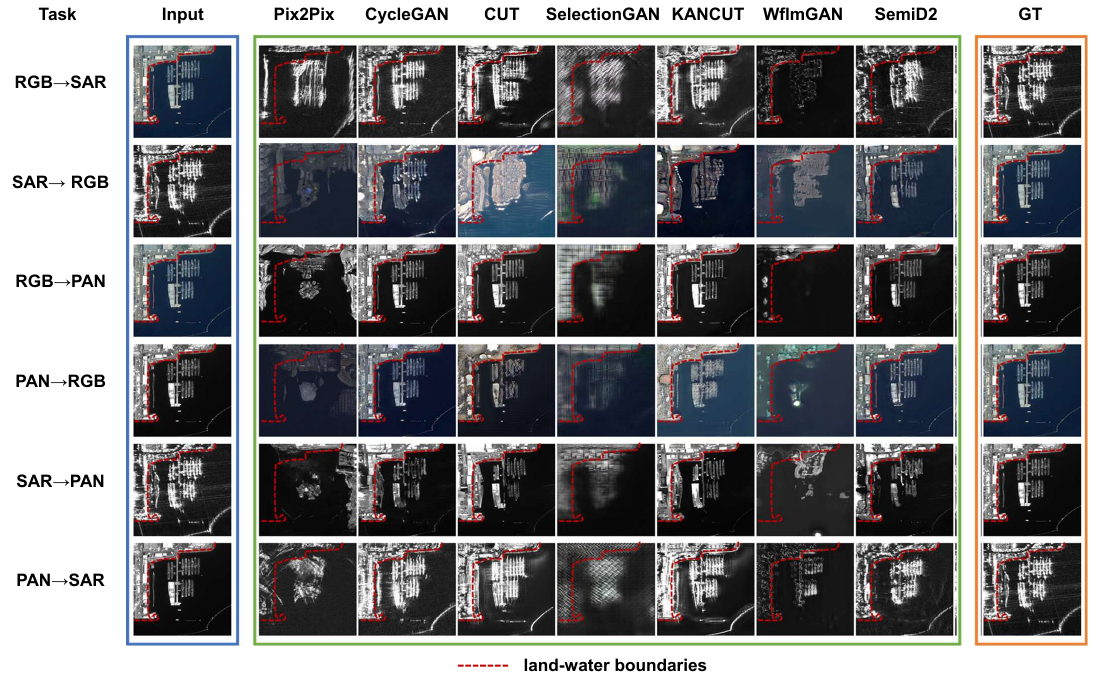}
    \caption{Task 3 Cross-Modal Generation - qualitative comparison between RGB, SAR, and PAN modalities. Each row shows input images and generation results from seven different methods, compared against the ground truth (GT).}
    \label{fig_006}
\end{figure*}

We comprehensively evaluate seven cross-modal generation methods on the SMART-Ship dataset. Table \ref{tab:multimodal_comparison} presents the quantitative results across different generation tasks, while Figure \ref{fig_006} provides visual comparisons that reveal important qualitative insights.

\textbf{Chronological and Domain-Specific Performance Analysis.}
Our benchmark spans 2017-2024 methods, revealing performance doesn't follow chronological progression. CycleGAN~\cite{CycleGAN2017} and Pix2Pix~\cite{Pix2Pix2017} (2017) show distinct strengths: CycleGAN achieves best average FID scores, indicating superior perceptual quality despite lower PSNR/SSIM. Figure \ref{fig_006} confirms CycleGAN preserves semantic structures like land-water boundaries that higher PSNR/SSIM methods misrepresent. Pix2Pix generates the target-domain images with more local textures but wrong global structures, which coincide with better PSNR but worse FID scores than CycleGAN. CUT~\cite{CUT2020} (2020) and KANCUT~\cite{KANCUT2024} (2024) excel in FID metrics for RGB$\rightarrow$PAN and SAR$\rightarrow$PAN tasks but score lower on reference-based metrics. SelectionGAN~\cite{SelectionGAN2022} produces poor visual results with excessive blurring despite competitive PSNR values, explaining its poor FID scores and highlighting limitations of reference-based metrics alone.

Domain-specific methods excel in pixel-level metrics. WflmGAN~\cite{WflmGAN2022} and SemiD2~\cite{SemiD2022} lead in average PSNR/SSIM, with WflmGAN achieving 13.88/0.335, 14.14/0.456, and 13.30/0.329 for RGB$\leftrightarrow$SAR, RGB$\leftrightarrow$PAN, and SAR$\leftrightarrow$PAN generation, respectively. SemiD2 achieves the highest SSIM (0.238) for RGB$\rightarrow$SAR tasks. 
However, these RS-specific methods show poor FID performance due to task-domain misalignment. Designed for terrestrial RGB$\leftrightarrow$SAR translation, they lack optimization for other modalities and adaptation to port and aquatic scenarios, deviating from typical urban/rural training environments. WflmGAN exhibits significant discrepancy between numerical performance and visual quality. Despite superior PSNR/SSIM, it produces severe semantic distortions, particularly at land-water boundaries. Figure \ref{fig_006} demonstrates WflmGAN incorrectly generating water bodies replacing land areas, altering scene semantics. This stems from its wavelet-based architecture prioritizing texture matching over semantic consistency, yielding high pixel-level similarity but poor semantic fidelity (high FID). Conversely, the dual discriminator design of SemiD2 better preserves land cover distinctions, although it does not match the semantic fidelity of CycleGAN.

\textbf{Task-Specific Method Effectiveness.}
Task difficulty analysis reveals a clear hierarchy correlated with cross-modal disparities. Using CycleGAN as a representative baseline, we observe distinct difficulty rankings: RGB$\leftrightarrow$PAN achieves the lowest FID score (73.37) due to shared optical sensing principles with only spectral differences; SAR$\leftrightarrow$PAN demonstrates intermediate complexity (FID: 132.04) where single-channel representations reduce spectral complexity; and RGB$\leftrightarrow$SAR proves most challenging (FID: 160.17), representing over double the perceptual distortion of RGB$\leftrightarrow$PAN. This substantial performance gap reflects fundamental differences between optical and radar sensing mechanisms, creating significant domain gaps challenging semantic preservation. The hierarchy (RGB$\leftrightarrow$PAN $<$ SAR$\leftrightarrow$PAN $<$ RGB$\leftrightarrow$SAR) is further validated by SSIM patterns, where RGB$\leftrightarrow$SAR exhibits the most pronounced divergence between pixel-level metrics (0.169) and perceptual quality, underscoring how modality gap magnitude directly impacts translation effectiveness.

\textbf{Metric Trade-off Analysis.}
Our results highlight a trade-off between reference-based metrics (PSNR/SSIM) and perceptual quality (FID), confirmed by visual assessment. High PSNR/SSIM methods (e.g., WflmGAN, SemiD2) show poor FID, while top FID methods (e.g., CycleGAN, CUT, KANCUT) underperform on PSNR/SSIM. Visual examination reveals the underlying mechanism:
High PSNR/SSIM methods ensure pixel-wise accuracy to some extent but may introduce semantic errors such as land being misclassified as water. Similarly, High FID methods preserve semantic structure and scene appearance to a certain degree, retaining key features like land-water boundaries despite pixel-level deviations.
This trade-off impacts RS applications with varying quality needs. Pixel-level tasks (e.g., change detection) favor high PSNR/SSIM, while visual interpretation benefits from better FID. Semantic correctness often outweighs pixel accuracy, making FID more relevant for practical RS use cases.

\subsection{Task 4: Pan-Sharpening}
\label{subsec:pansharpening}

\begin{table*}[t]
\caption{Quantitative comparison of different methods on pan-sharpening and cross-modal change detection task.}
\begin{subtable}[t]{0.5\textwidth}
\centering
\caption{Task 4: Pan-Sharpening.}
\label{tab:pan_sharpening}
\resizebox{0.9\textwidth}{!}{%
\begin{tabular}{l|c|c|c|c|c|c}
\toprule
\multirow{2}{*}{\textbf{Method}} & \multirow{2}{*}{\textbf{Year}} & \multirow{2}{*}{\textbf{Params}} & \multicolumn{4}{c}{\textbf{Metrics}} \\
\cmidrule(lr){4-7}
 & & & \textbf{PSNR}$\uparrow$ & \textbf{SSIM}$\uparrow$ & \textbf{ERGAS}$\downarrow$ & \textbf{SAM}$\downarrow$ \\
\midrule
PanNet~\cite{Yang_2017_ICCV} & 2017 & 0.151M & 18.20& 0.631& 5.945& 16.529 \\
MSDCNN~\cite{yuan2018multiscale} & 2018 & 0.18M & \underline{23.37}& \underline{0.811}& \underline{3.997}& \underline{5.428} \\
DiCNN~\cite{he2019dicnn} & 2019 & 0.041M & 21.46& 0.778& 4.560& 6.587 \\
BDPN~\cite{zhang2019pan} & 2019 & 2.95M & 19.88& 0.744& 5.085& 7.927 \\
FusionNet~\cite{deng2020detail} & 2020 & 0.149M & 22.31&  0.801& 4.151& 7.041 \\
PNN~\cite{Ciotola2022} & 2022 & 0.074M &   19.81& 0.736& 4.882& 9.094 \\
LAGNet~\cite{jin2022lagconv} & 2022 & 0.147M & \textbf{25.94}& \textbf{0.839}&  \textbf{3.757}& \textbf{4.496} \\
\bottomrule
\end{tabular}%
}
\end{subtable}%
\hfill
\begin{subtable}[t]{0.48\textwidth}
\centering
\caption{Task 5: Cross-Modal Change Detection.
}
\label{tab:change_detection}
\resizebox{0.98\textwidth}{!}{%
\begin{tabular}{l|c|c|c|c|c|c}
\toprule
\multirow{2}{*}{\textbf{Method}} & \multirow{2}{*}{\textbf{backbone}} & \multirow{2}{*}{\textbf{Params}}& \multicolumn{4}{c}{\textbf{Metrics}} \\
\cmidrule(lr){4-7}
 & & & \textbf{mF1}$\uparrow$ & \textbf{P}$\uparrow$ & \textbf{R}$\uparrow$ & \textbf{mIoU}$\uparrow$ \\
\midrule
SS-BITMIX~\cite{chen2021bit} & Resnet18 & 0.59M& 63.7 & 67.9 & 61.2 & 57.7\\
SS-ChangeFormer~\cite{cf} & - & 0.74M& 62.1 & 68.3 & 59.0 & 56.6\\
SS-Segb0~\cite{xie2021segformer} & SegFormer-b0 & 0.74M& 67.9 & 75.5 & 63.8 & 60.7\\
SS-Segb1~\cite{xie2021segformer} & SegFormer-b1 & 2.73M& \textbf{71.5} & \textbf{80.3} & \underline{66.7} & \textbf{63.5}\\
SS-Segb2~\cite{xie2021segformer} & SegFormer-b2 & 4.94M& \underline{71.1} & \underline{72.9} & \textbf{69.5} & \underline{63.2}\\
\bottomrule
\end{tabular}%
}
\vspace{0.5ex}

\parbox{0.98\textwidth}{\footnotesize
Note: All methods are prefixed with \texttt{"}SS\texttt{"}, indicating benchmark modifications for our cross-modal change detection task.
}
\end{subtable}
\end{table*}

\begin{figure*}[t]
    \centering
    \includegraphics[width=0.9\linewidth]{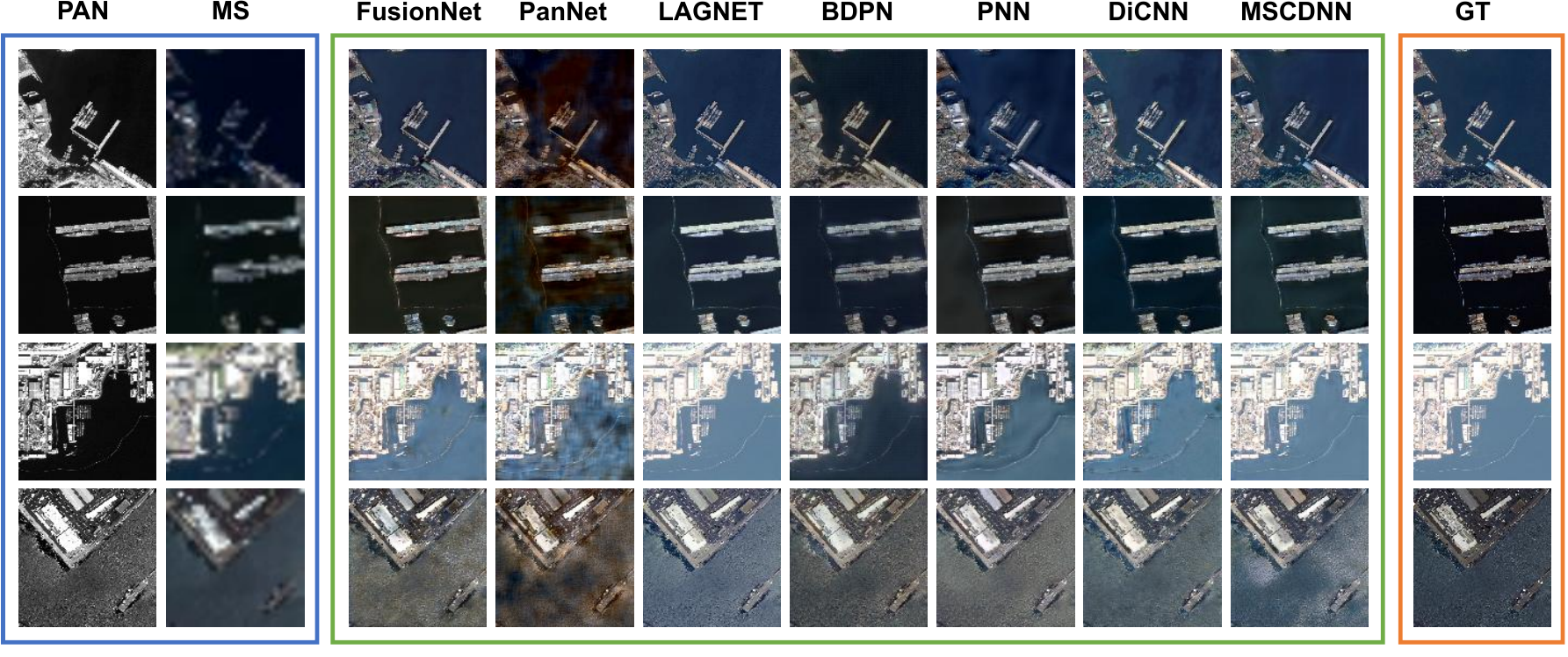}
    \caption{Task 4: Pan-Sharpening - qualitative comparison. Each row shows input images, paired of high resolution PAN and low resolution MS, and RGB generated results from seven different methods, compared against the ground truth (GT).}
    \label{fig_007}
\end{figure*}

Pan-sharpening facilitates the fusion of complementary satellite imagery modalities, offering significant advantages for enhanced spatial-spectral resolution in RS applications. This study evaluates multiple fusion approaches, encompassing trending deep learning architectures for integrating PAN and MS imagery. The PAN$\leftrightarrow$MS fusion addresses the fundamental trade-off between spatial and spectral resolution inherent in satellite sensor design, enabling the synthesis of imagery that preserves both fine spatial details and rich spectral characteristics.

\subsubsection{Task Formulation}
Given an input high-resolution panchromatic image $x^{m_\mathrm{pan}} \in \mathcal{X}^{m_\mathrm{pan}(h \times w \times 1)}$ and a low-resolution multi-spectral image $x^{m_\mathrm{ms}} \in \mathcal{X}^{m_\mathrm{ms}(h/r \times w/r \times c)}$, where $m_\mathrm{pan}, m_\mathrm{ms} \in \mathcal{M}$ denote the panchromatic and multi-spectral modalities, $r$ is the downsampling ratio, and $c$ is the number of spectral bands, pan-sharpening aims to synthesize a high-resolution RGB image $\hat{x}^{m_\mathrm{rgb}} \in \mathcal{X}^{m_\mathrm{pan}(h \times w \times 3)}$. This fusion task can be formalized as learning an optimal mapping $S: \mathcal{X}^{m_\mathrm{rgb}} \times \mathcal{X}^{m_\mathrm{ms}} \rightarrow \mathcal{X}^{m_\mathrm{rgb}}$, where $\mathcal{X}^{m}$ denotes the image domain of modality $m$:
\begin{equation}
\hat{x}^{m_\mathrm{rgb}} = S\left(x^{m_\mathrm{pan}}, x^{m_\mathrm{ms}}\right)
\end{equation}
The objective is to preserve spatial texture from $x^{m_\mathrm{pan}}$ while maintaining accurate spectral characteristics in the reconstructed RGB image.

\subsubsection{Evaluation Setting}
\label{subsec:pansharpening_setting}
\textbf{Dataset.}
Experiments are conducted on a pan-sharpening dataset derived as a subset from SMART-Ship, following the original train-test split settings. Each triplet contains spatially aligned imagery of the same geographical location across three components: high-resolution PAN (single channel), low-resolution MS (3 channels), and high-resolution RGB (3 channels, serving as ground truth).

\textbf{Implementation Details.} We implement and evaluate seven representative pan-sharpening methods: BDPN~\cite{zhang2019pan}, MSDCNN~\cite{yuan2018multiscale}, PNN~\cite{Ciotola2022}, DiCNN~\cite{he2019dicnn}, FusionNet~\cite{deng2020detail}, PanNet~\cite{Yang_2017_ICCV}, and LAGNet~\cite{jin2022lagconv}. All implementations are based on the same PyTorch training framework and trained on GPU platform.
When reproducing benchmark methods, the original method input exhibits varying input resolutions, with PAN images ranging from 64×64 to 256×256 pixels and MS images ranging from 16×16 to 64×64 pixels. For fair comparison and computational resource balance, all PAN and MS images are resized to standardized dimensions of 128×128 and 32×32 pixels, respectively. For each model, we conduct comprehensive experiments on the pan-sharpening task, taking high-resolution PAN (128×128) and low-resolution MS (32×32) images as inputs to generate high-resolution RGB images (128×128). To ensure fair comparison, all models are evaluated using the same dataset partition and preprocessing pipeline.

All models are trained for 500 epochs using the Adam optimizer with an initial learning rate of 0.0001. The training workflow alternates between 25 training steps and one testing step for continuous performance monitoring. The batch size is 16 across all experiments, with 4 data loaders per GPU to ensure efficient training. Model checkpoints are saved every 50 epochs, and training logs are recorded at the same interval.

\textbf{Evaluation Metrics.} 
The fused RGB image is evaluated using four complementary indicators: PSNR, SSIM, Error Relative Global Dimensionless Synthesis (ERGAS)~\cite{panqual}, Spectral Angle Mapper(SAM)\cite{panqual}.

\subsubsection{Experimental Results}
\textbf{Overall Performance Assessment.} 
As shown in Table~\ref{tab:pan_sharpening}, our evaluation indicates that LAGNet achieves the best performance with PSNR of 25.94 dB and SSIM of 0.839, effectively preserving both spatial details and spectral fidelity. The second-tier methods show consistently competitive performance across evaluation metrics, establishing a clear performance hierarchy that distinguishes advanced approaches from conventional techniques. 
While all evaluated methods demonstrate certain effectiveness in spatial-spectral fusion tasks, which is visualized in Figure~\ref{fig_007}, there remains considerable room for performance enhancement across the field. Overall, our findings highlight the critical importance of domain-specific architectural design in pan-sharpening effectiveness, suggesting that our SMART-ship dataset presents significant opportunities for further methodological advancement.

\subsection{Task 5: Cross-Modal Change Detection}
\label{subsec:changedetection}
Cross-modal change detection facilitates the identification of temporal variations across heterogeneous modalities, bridging optical and radar sensing domains for weather-independent monitoring in RS applications. 
The three implemented frameworks (another two with more stacked blocks but similar structure), named SS-BITMIX, SS-ChangeFormer, and SS-SegFormer, are adapted from the widely-used BIT~\cite{chen2021bit}, ChangeFormer~\cite{cf}, and SegFormer~\cite{xie2021segformer} architectures, respectively. We introduce task-specific modifications and engineering optimizations for each model to enable robust cross-modal change detection on optical and SAR data; these models are denoted with the \texttt{"}SS\texttt{"} prefix to indicate our specialized adaptation for the considered benchmark.

\begin{figure*}
    \centering    \includegraphics[width=0.9\linewidth]{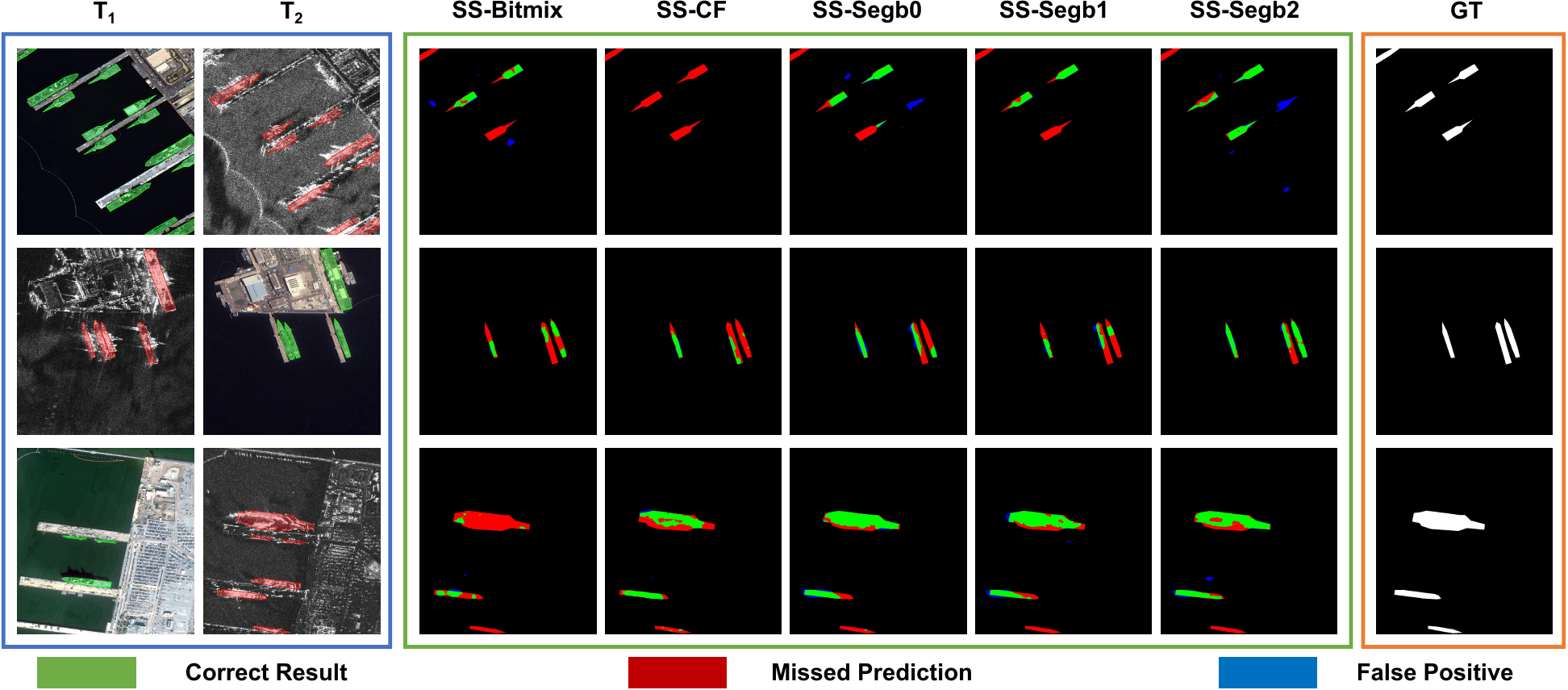}
    \caption{Task 5: Cross-Modal Change Detection - qualitative comparison. Bi-temporal RGB-SAR image pairs and change predictions compared to ground truth.
    }
    \label{fig_008}
\end{figure*}

\subsubsection{Task Formulation}
Given a bi-temporal image pair consisting of an RGB image $x^{m_\mathrm{rgb}} \in \mathcal{X}^{m_\mathrm{rgb}(h \times w \times 3)}$ from time $t_{RGB}$and a SAR image $x^{m_\mathrm{sar}} \in \mathcal{X}^{m_\mathrm{sar}(h \times w \times 1)}$ from time $t_{SAR}$, cross-modal change detection aims to identify and localize changes between different time periods, producing a binary change mask $\hat{B} \in \{0,1\}^{h \times w}$ where 1 indicates changed regions and 0 indicates unchanged areas. This change detection task can be formalized as learning an optimal mapping function $C: \mathcal{X}^{m_\mathrm{rgb}} \times \mathcal{X}^{m_\mathrm{sar}} \rightarrow \mathcal{X}^{mask}$ such that:
\begin{equation}
\hat{x}^{mask} = C(x^{m_\mathrm{rgb}}, x^{m_\mathrm{sar}})
\end{equation}
The framework addresses the challenge of cross-modal temporal analysis by learning robust feature representations that can effectively compare information across different imaging modalities (optical RGB vs. microwave SAR) while maintaining sensitivity to meaningful land cover changes.

\subsubsection{Evaluation Setting}
\label{subsec:cross_modal change detection}
\textbf{Dataset.}
We adopt a subset of the SMART-Ship dataset specifically curated for cross-modal change detection tasks, replicating the data partitioning strategy from prior investigations to ensure methodological consistency.
Each pair contains spatially aligned imagery of the same geographical location across RGB and SAR images.
We use the dataset with binary change labels that distinguish between the \texttt{"}change\texttt{"} and \texttt{"}no-change\texttt{"} regions throughout the scenes. This subset provides comprehensive coverage of various change scenarios in maritime environments.
The CD subset of SMART-Ship maintains spatial consistency with image pairs captured at different time points, ensuring robust evaluation of cross-modal change detection algorithms under varying atmospheric and sea conditions typical of maritime surveillance scenarios. Specifically, we introduced 1292 changed instances across the train and test splits, providing sufficient changed samples.

\textbf{Implementation Details.}
All models are trained for 200,000 iterations using the Adamw optimizer with an initial learning rate of 0.0001 and optimized by cross-entropy loss. The training workflow alternates between 8000 training steps and one testing step for continuous performance monitoring. To ensure efficient training, the batch size is set to 8 for all experiments, with four dataloaders per GPU. Model checkpoints are saved every 8000 iterations, and training logs are recorded at the same interval.

For detailed model structure implementation, we adopt different pretrained backbones as shown in Table~\ref{tab:change_detection}. Models with a specified backbone inherit the pretrained weights and continue training on our dataset, while models without a backbone entry are trained from scratch. Due to the lack of open-source implementations and unified benchmarks in many cross-modal change detection works, we adapt excellent methods from homogeneous change detection to validate the SMART-Ship dataset. Specifically, we adopt a dual-branch design for the feature extraction backbone, constructing separate RGB and SAR branches with independent parameters instead of shared ones. We employ a multi‑layer perceptron (MLP) to map SAR inputs into 3-channel format.

\textbf{Evaluation Metrics.}
We evaluate change detection results with four complementary indicators\cite{ Changen2, fcd, iou}: mean F1-score (mF1), precision (P), recall (R), and mean IoU (mIoU).

\subsubsection{Experimental Results}
\textbf{Overall Performance Assessment and Architecture-Specific Analysis.}
Our proposed SS-Seg framework validates the feasibility of cross-modal change detection tasks and achieves certain effectiveness through architectural adaptations for RGB$\leftrightarrow$SAR temporal analysis. SS-Seg with SegFormer-B1 backbone performs best in the evaluation, which is shown in Figure~\ref{fig_008}, achieving mF1 of 71.5\%, precision of 80.3\%, recall of 66.7\%, and mIoU of 63.5\%, demonstrating the effectiveness of the framework in cross-modal feature alignment and temporal change localization. SS-Seg with SegFormer-B2 also achieves favorable results, while the hybrid approach SS-BITMIX reaches 63.72\% mF1, though results indicate that mixed pretrained convolutional backbones have certain limitations in cross-modal representational capacity compared to pure transformer architectures.

\section{Discussion}
\label{sec:analysis}
\subsection{Advantages of Multi-modal Data over Unimodal Data}
The utilization of multi-modal data enhances the long-term maritime surveillance. This enables the robust scene interpretation through the collaboration of different sensors. As demonstrated in Figure~\ref{fig_004}, the combination of RGB and SAR images allows for accurate ship detection under challenging conditions such as cloud occlusion or low visibility, where RGB-only and SAR-only solutions fail to detect and distinguish the ships. Besides, the complementary information between different modalities like PAN and MS could reconstruct the high-fidelity scene details and lead to better detection performance in Table~\ref{tab:detection} and ~\ref{tab:pan_sharpening}.
Moreover, to bridge the discrepancy between various modalities could learn the shared semantic representations of the scenario, which supports the cross-modal tasks such as instance re-identification, image generation, and change detection for long-term maritime monitoring.

\subsection{Challenges on Multi-modal Analysis of Maritime Scenarios}
The fundamental difference in imaging mechanisms, particularly between optical and SAR modalities, poses a major challenge for effective multi-modal analysis. The structural diversity of maritime objects and the dynamic nature of maritime environments further impose burdens on multi-modal alignment and fusion. This challenge is evident in our benchmarks, where cross-modal image generation between RGB and SAR yields only marginal improvements, as reported in Table~\ref{tab:multimodal_comparison}. Moreover, real-world maritime object datasets like SMART-ship inevitably suffer from the long-tailed category distributions and significant scale variance, leading to performance biases in detection and re-identification tasks. Future exploration of these tasks could emphasize consistent performance across frequent and rare object categories.

\subsection{Future Work}
Extensive evaluations on the SMART-Ship benchmark open up promising opportunities for advancing multi-modal maritime RS interpretation. Beyond validating the effectiveness of existing fusion strategies, it highlights the need for modality-specific architectures since current backbones struggle to extract sufficiently distinctive feature representations of each modality. Incorporating lightweight modality-adaptive modules could further enhance the generalization and efficiency of multi-modal analysis across diverse satellite platforms. Moreover, the rich annotations in the SMART-Ship dataset inspire the joint exploration of multiple downstream tasks, like object detection, semantic segmentation, and change detection, for comprehensive understanding of maritime scenes.

\section{Conclusion}
\label{sec:conclusion}
In this paper, we introduce the SMART-Ship, a unified multi-modal ship dataset with fine-grained annotations for a wide range of RS interpretation tasks.
The dataset provides 1092 multi-modal image sets with spatial registration and temporal synchronization across five key modalities: visible-light, SAR, panchromatic, multi-spectral, and near-infrared. Ship instances in each set are annotated with polygonal location information, fine-grained categories, instance-level identifiers, and change region masks, organized hierarchically to support multiple maritime scene interpretation tasks. Moreover, we have established benchmarks for five downstream tasks: multi-modal detection, cross-modal re-identification, cross-modal generation, pan-sharpening, and cross-modal change detection. Extensive evaluations of representative methods demonstrate the effectiveness of the proposed dataset in terms of object diversity, scene typicality, and task relevance. Experimental results reveal both the challenges and opportunities of multi-modal analysis in maritime scenarios. We expect the SMART-Ship dataset to become a valuable resource for the remote sensing community, promoting future research on multi-modal fusion, cross-modal learning, and multi-task collaboration.

\bibliographystyle{IEEEtran}
\bibliography{dataset_paper}

 
\vspace{11pt}

\begin{IEEEbiographynophoto}{Chen-Chen Fan} (Member, IEEE) is currently a Postdoctoral Researcher with the Department of Electronic Engineering, Tsinghua University, Beijing, China. He received the B.E. degree from Beijing University of Chemical Technology, Beijing, China, in 2018, and the Ph.D. degree from the Institute of Automation, Chinese Academy of Sciences and University of Chinese Academy of Sciences, Beijing, China, in 2023. His research interests include computer vision, multimodal data processing, and remote sensing.
\end{IEEEbiographynophoto}
\vspace{-11mm}
\begin{IEEEbiographynophoto}{Peiyao Guo}(Member, IEEE) is a Postdoctoral Researcher in the Department of Electronic Engineering, Tsinghua University, Beijing, China. She received the B.E., M.S., and Ph.D. degrees in Electronic Science and Engineering from Nanjing University, Jiangsu, China, in 2016, 2019, and 2024, respectively. Her current research focuses on image/video processing, computational imaging, and multi-modal fusion.
\end{IEEEbiographynophoto}
\vspace{-11mm}
\begin{IEEEbiographynophoto}{Linping Zhang} (Student Member, IEEE) received the B.S. degree from Southeast University, Nanjing, China, in 2022. He is currently pursuing the Ph.D. degree with Tsinghua University, Beijing, China. His research interests include information fusion, object re-identification, object detection, and remote sensing.
\end{IEEEbiographynophoto}
\vspace{-11mm}
\begin{IEEEbiographynophoto}{Kehan Qi} received the B.S. degree from Harbin Institute of Technology, Harbin, China, in 2024. He is currently pursuing the Ph.D. degree with Shenzhen International Graduate School, Tsinghua University, Shenzhen, China. His research interests include remote sensing, multimodal fusion, and computer vision.
\end{IEEEbiographynophoto}
\vspace{-11mm}
\begin{IEEEbiographynophoto}{Haolin Huang} received the B.S. degree from Jilin University, Changchun, China, in 2022. He is currently pursuing the M.S. degree with Tsinghua Shenzhen International Graduate School, Shenzhen, China. His research interests include information fusion, change detection, and remote sensing.
\end{IEEEbiographynophoto}
\vspace{-11mm}
\begin{IEEEbiographynophoto}{Yong‑Qiang Mao} (Member, IEEE) received the B.Sc. degree from Wuhan University, Wuhan, China, in 2019, and the Ph.D. degree from Aerospace Information Research Institute, Chinese Academy of Sciences and University of Chinese Academy of Sciences, Beijing, China, in 2024. He is currently a postdoctoral researcher at the Department of Electronic Engineering, Tsinghua University. His research interests include remote sensing, computer vision, pattern recognition, especially multimodal fusion, 3D computer vision, and large foundation models.
\end{IEEEbiographynophoto}
\vspace{-11mm}
\begin{IEEEbiographynophoto}{Yuxi Suo} received the B.E. degree from University of Chinese Academy of Sciences, Beijing, China, in 2020, and the Ph.D. degree with the Aerospace Information Research Institute, Chinese Academy of Sciences, Beijing, China, in 2025. He is currently an Assistant Professor with the Department of Electronic Engineering, Tsinghua University. His research interests include deep learning, remote sensing image interpretation, and the improvement of spaceborne SAR image quality.
\end{IEEEbiographynophoto}
\vspace{-11mm}
\begin{IEEEbiographynophoto}{Zhizhuo Jiang} (Member, IEEE) received the B.S. and M.S. degrees in electronic engineering from Harbin Institute of Technology, Harbin, China, in 2014 and 2016, respectively. He received the Ph.D. degree in electronic engineering from Tsinghua University, Beijing, China, in 2022. He is currently a Post-Doctoral Fellow with the Shenzhen International Graduate School, Tsinghua University, Shenzhen, China. His main research interests include cross-domain information fusion, multimodal data processing, and fine-grained classification.
\end{IEEEbiographynophoto}
\vspace{-11mm}
\begin{IEEEbiographynophoto}{Yu Liu} (Member, IEEE) received the B.S. and Ph.D. degrees in information and communication engineering from Naval Aviation University, Yantai, China, in 2008 and 2014, respectively. Since 2014, he has been with the faculty of Naval Aviation University, where he is currently a Professor with the Research Institute of Information Fusion. From 2016 to 2018, he was a Post-Doctoral Researcher with the Department of Information and Communication Engineering, Beihang University, Beijing, China. He is currently a Research Fellow in the Department of Electronics at Tsinghua University. His research interests include multi-sensor fusion and remote data processing.
\end{IEEEbiographynophoto}
\vspace{-11mm}
\begin{IEEEbiographynophoto}{You He} (Member, IEEE) received the Ph.D. degree in electronic engineering from Tsinghua University, Beijing, China, in 1997. He is currently a Professor at Tsinghua University. He has published over 300 academic articles. He is the author of Radar Target Detection and CFAR Processing (Tsinghua University Press), Multi-sensor Information Fusion with Applications, and Radar Data Processing with Applications (Publishing House of Electronics Industry). His current research interests include detection and estimation theory, CFAR processing, distributed detection theory, and multi-sensor information fusion. Prof. He is a Fellow Member of the Chinese Academy of Engineering. In 2017, he won the top prize in science and technology of Shandong Province.
\end{IEEEbiographynophoto}
\vfill

\end{document}